\documentclass[pmlr]{jmlr}

\RequirePackage{graphicx}
 \usepackage{booktabs}
\usepackage{longtable}
\makeatletter
\def\set@curr@file#1{\def\@curr@file{#1}} 
\makeatother
\usepackage[load-configurations=version-1]{siunitx} 

\theorembodyfont{\upshape}
\theoremheaderfont{\scshape}
\theorempostheader{:}
\theoremsep{\newline}

\usepackage{colortbl}
\definecolor{lightgray}{rgb}{0.9, 0.9, 0.9}
\definecolor{cluster2}{HTML}{ffd500}
\definecolor{cluster3}{HTML}{009eff}
\definecolor{cluster4}{HTML}{ff00ed}

\usepackage{array}
\usepackage{multirow} 
\usepackage{float}
\usepackage{dcolumn}        
\usepackage{algorithm}
\usepackage{listings}

\newcolumntype{C}{D{,}{,}{-1}} 
\newcolumntype{d}[1]{D{,}{,}{#1}}
\newcolumntype{.}{D{.}{.}{-1}}
\newcolumntype{,}{D{|}{|}{-1}}
\newcommand\mc[1]{\multicolumn{1}{c}{#1}} 

\newcommand{\bs}[1]{\boldsymbol{#1}}

\newcommand{\RR}[1]{\mathbb{R}^{#1}}

\newcommand{\equal}[1]{{\hypersetup{linkcolor=black}\thanks{#1}}}

\title[Phenotype Clustering of Disease Trajectories]{Clustering of Disease Trajectories with Explainable Machine Learning: A Case Study on Postoperative Delirium Phenotypes}

\author{\Name{Xiaochen Zheng}\equal{Correspondence to Xiaochen Zheng.}
       \Email{xiaochen.zheng@uzh.ch}\\ 
       \addr University of Zürich\\
       \vspace{-3.6mm}
       \AND
       \Name{Manuel Schürch}
       \Email{manuel.schuerch@uzh.ch}\\ 
       \addr University of Zürich\\
       \vspace{-3.6mm}
       \AND
       \Name{Xingyu Chen}
       \Email{xingyuchen@ethz.ch}\\ 
       \addr ETH Zürich\\
       \vspace{-3.6mm}
       \AND
       \Name{Maria Angeliki Komninou}
       \Email{mariaangeliki.komninou@usz.ch}\\ 
       \addr University Hospital Zürich\\
       \vspace{-3.6mm}
       \AND
       \Name{Reto Schüpbach}
       \Email{reto.schuepbach@usz.ch}\\ 
       \addr University Hospital Zürich\\
       \vspace{-3.6mm}
       \AND
       \Name{Ahmed Allam}
       \Email{ahmed.allam@uzh.ch}\\ 
       \addr University of Zürich\\
       \vspace{-3.6mm}
       \AND
       \Name{Jan Bartussek}\equal{Equal senior authorship.}
       \Email{jan.bartussek@usz.ch}\\ 
       \addr University Hospital Zürich\\
       \vspace{-3.6mm}
       \AND
       \Name{Michael Krauthammer}\footnotemark[2]
       \Email{michael.krauthammer@uzh.ch}\\ 
       \addr University of Zürich, University Hospital Zürich\\
       \vspace{-10mm}
       } 

\begin{document}

\maketitle

\begin{abstract}
The identification of phenotypes within complex diseases or syndromes is a fundamental component of precision medicine, which aims to adapt healthcare to individual patient characteristics. Postoperative delirium (POD) is a complex neuropsychiatric condition with significant heterogeneity in its clinical manifestations and underlying pathophysiology. We hypothesize that POD comprises several distinct phenotypes, which cannot be directly observed in clinical practice. Identifying these phenotypes could enhance our understanding of POD pathogenesis and facilitate the development of targeted prevention and treatment strategies. In this paper, we propose an approach that combines supervised machine learning for personalized POD risk prediction with unsupervised clustering techniques to uncover potential POD phenotypes. We first demonstrate our approach using synthetic data, where we simulate patient cohorts with predefined phenotypes based on distinct sets of informative features. We aim to mimic any clinical disease with our synthetic data generation method. By training a predictive model and applying SHapley Additive exPlanations (SHAP), we show that clustering patients in the SHAP feature importance space successfully recovers the true underlying phenotypes, outperforming clustering in the raw feature space. We then present a case study using real-world data from a cohort of elderly surgical patients. We train machine learning models on multimodal electronic health record data that span the preoperative, intraoperative and postoperative periods to predict personalized POD risk. Subsequent clustering of patients based on their SHAP feature importances reveals distinct subgroups with differing clinical characteristics and risk profiles, potentially representing POD phenotypes. These results showcase the utility of our approach in uncovering clinically relevant subtypes of complex disorders like POD, paving the way for more precise and personalized treatment strategies.
\end{abstract}

\section{Introduction}
The identification of phenotypes in complex diseases and syndromes is essential for precision medicine, which seeks to customize healthcare based on the specific traits of individual patients. This process is not only a foundation of patient-centered treatment but also paves the way for more refined, tailored therapeutic interventions. By focusing on these unique characteristics, precision medicine enhances the efficacy of healthcare delivery, making it more responsive to the diverse needs of patients. A distinct example of the importance of phenotyping is observed in the study of sepsis ~\citep{sepsis_phenotype}, a heterogeneous syndrome where the identification of distinct clinical phenotypes has the potential to enable more tailored therapy and enhance patient care outcomes. In this retrospective analysis of data sets from patients with sepsis, four distinct clinical phenotypes were identified ($\alpha$, $\beta$, $\gamma$, and $\delta$), each correlating with specific host-response patterns and clinical outcomes. This differentiation highlights how clinical manifestations of sepsis can be linked to underlying biological pathways and patient variability. 
In general, the recognition of distinct clinical and biological phenotypes is enhancing our comprehension of how clinical manifestations correlate with underlying pathways and variability among patients ~\citep{delirium_subgroup4} and is critical for several reasons: it facilitates the development of personalized treatment, improves diagnostic accuracy, and optimizes patient careoutcomes~\citep{morley2021phenotypic,niemann2024heterogeneity}.

Delirium is a serious neuropsychiatric postoperative complication that occurs in up to 46\% of the general surgical population~\citep{whitlock2011postoperative}. Symptoms of postoperative delirium include a rapid onset of confusion, attention deficits, disorganized thinking, and fluctuating levels of consciousness, alongside memory issues, mood swings, behavioral changes, and sleep disturbances, highlighting the need for prompt recognition and effective treatment~\citep{vijayakumar2014postoperative}. Untreated, it significantly raises distress, mortality rates, and the risk of long-term cognitive decline~\citep{milbrandt2004costsdelirium,field2013delirium,wilson2020delirium, komninou2024smoking}. Therefore, efficient treatment or prevention holds a key to improving clinical management through early detection and the development of effective treatment strategies~\citep{fong2009deliriumelder}, which in turn can reduce healthcare costs associated with prolonged hospital stays. 

Being a complex syndrome, delirium presents challenges in understanding its underlying mechanisms. Its common occurrence in ICU and postoperative environments does not translate to a clear understanding of its pathophysiology~\citep{gunther2008pathophysiology}. The prevailing hypothesis suggests that delirium occurs from disruptions in neurotransmitter balance, influenced by certain illnesses, neuroinflammatory responses, or medical treatments making its treatment and prevention complicated. Subsequently, understanding its neurobiological mechanism could provide crucial insights into brain function under stress and illness, which might shed light on other neuropsychiatric and neurological disorders~\citep{van2012development, lindroth2018systematic}.

Personalized delirium risk prediction using machine learning (ML) algorithms, based on comprehensive perioperative patient trajectory data, exhibits great potential in this direction. Supervised ML methods have proven to successfully provide significant capability in predicting risks on an individual basis~\citep{delirium_ml1}. Accurate prediction of the likelihood of delirium before and after surgery enables healthcare providers to implement early interventions tailored to particular patient needs~\citep{fong2009deliriumelder}. Nonetheless, within a heterogeneous patient group, individuals may present identical risk levels but differ in their disease trajectory and development of delirium, which require different interventions~\citep{subgroup4}. This highlights a critical gap in translating pure ML risk prediction approaches into clinical routine and personalized intervention strategies.

Our main hypothesis is that postoperative delirium (POD) has several phenotypes that can be identified through data-driven approaches. To address this hypothesis, we propose a two-fold approach. First, we develop a synthetic case study to demonstrate the feasibility of our method in identifying phenotypes within a controlled environment. Second, we apply our approach to a real-world case study of perioperative delirium to uncover potential phenotypes and gain insights into the underlying factors contributing to the development of delirium.

The proposed approach involves training perioperative prediction models for delirium, followed by the application of explainability techniques, such as SHapley Additive exPlanations (SHAP)~\citep{shap1,shap3,shap2}, to assess the significance of various features. By clustering patients based on these SHAP values, we aim to identify the key factors contributing to the development of delirium, thereby uncovering potential causes and reasons why certain patients are more susceptible to developing delirium. Additionally, this approach contributes in exploring and defining different subtypes of delirium based on learned features and corresponding importance, revealing distinct phenotypes characterized by distinct and influential clinical features for causing this condition. Several recent studies have proposed methods for clustering patient phenotypes using time-series data, such as representing patient trajectories in latent feature spaces~\citep{holland2023novelclustering1}, discovering predictive temporal patterns~\citep{aistatnovelclustering2}, learning patient representations through contrastive learning~\citep{noroozizadeh2023novelclustering3}, optimizing clustering performance with deep learning~\citep{lee2020novelclustering4}, simultaneously performing clustering and classification for risk prediction~\citep{srivastava2023novelclustering5}, and leveraging semi-supervised latent temporal processes with generative modeling~\citep{trottet2023generative}. However, none of these approaches performs phenotype clustering in the explanability space with multimodal data. 
\\
\\
Our contributions to this paper are:

\paragraph{Comprehensive ML Prediction:} We propose a robust ML approach for predicting the personalized risk of postoperative delirium leveraging the rich spectrum of multimodal electronic health record (EHR) data. We provide risk estimates that include the pre, intra and postoperative stages that can be used for the early detection of POD.

\paragraph{Personalized Explanations:} The holistic machine learning prediction model can be used to provide detailed average and personalized explanations (SHAP values) for the development of postoperative delirium in the form of important features, for the different stages of each feature in the patient's input vector. This approach can shed light on the individual feature contributions to the model's predictions.

\paragraph{Clustering of Phenotypes:} Based on the personalized and data-driven explanations of the predictive ML models at different stages, we present an unsupervised clustering approach, which enables the identification of distinct patient phenotypes and uncovers hidden phenotypes within the temporal development of POD, leading to gaining a better understanding, and allowing tailored and more personalized interventions of POD.

\subsection*{Generalizable Insights about Machine Learning in the Context of Healthcare}
First, we demonstrate how combining predictive modeling with post-hoc explainability techniques like SHAP values enables the identification of personalized risk factors and drivers of disease outcomes for individual patients. Second, we show that unsupervised clustering of patients based on their personalized feature importance scores is a powerful approach for uncovering hidden disease subtypes and phenotypes in a data-driven manner without requiring prior knowledge. Finally, our results highlight the potential of leveraging the full spectrum of multimodal clinical data across different perioperative stages with machine learning in order to generate holistic patient trajectory insights. Together, these generalizable techniques offer an exciting path forward for 
personalized healthcare.
\section{Methodology}\label{method}


\subsection{Hypothesis of Multiple Phenotypes in Cinical manifestation}
\label{hypothesis}
We begin our exploration by considering a cohort represented as $\mathcal{C}_N = \{\bs{x}_i, y_i\}_{i=1}^N$, where $\bs{x}_i \in \RR{D}$ denotes the set of features and $y_i \in \{0,1\}$ represents the associated labels within a clinical manifestation context, where $y_i = 1$ represents a positive manifestation. Within this cohort, we hypothesize the existence of multiple phenotypes that are not directly observable in clinical practice. To demonstrate our hypothesis, we will generate synthetic data for which we know the ground truth. In particular, we categorize the features of $\bs{x}_i$ into three distinct types: \texttt{shared}, \texttt{informative}, and \texttt{noisy}, as denoted as $\bs{x}_i = \{x_{i}^{\texttt{shared}}, x_{i}^{\texttt{informative}}, x_{i}^{\texttt{noisy}}\}$. Here, \texttt{shared} features refer to those common across all phenotypes, whereas \texttt{informative} features are unique and predominant within specific phenotypes, playing a crucial role in their differentiation. Both \texttt{shared} and \texttt{informative} features are instrumental in determining the phenotypes, while \texttt{noisy} features represent extraneous information that does not contribute to phenotype identification. This conceptualization allows us to define a phenotype through a binary-valued function depending on its \texttt{shared} and \texttt{informative} features, denoted as:
\begin{align}
\label{eq:phenotype}
\begin{split}
\text{ph}_{i,\alpha} &= f_{\alpha}(x_{i, \alpha}^{\texttt{informative}}) \\
\text{ph}_{i,\beta} &= f_{\beta}(x_{i}^{\texttt{shared}}, x_{i, \beta}^{\texttt{informative}}) \\
\text{ph}_{i,\gamma} &= f_{\gamma}(x_{i}^{\texttt{shared}}, x_{i, \gamma}^{\texttt{informative}}) \\
\text{ph}_{i,\delta} &= f_{\delta}(x_{i}^{\texttt{shared}}, x_{i, \delta}^{\texttt{informative}}) \\
\end{split}
\end{align} where function $f$ should return true if the data point belongs to the phenotype, and false otherwise.

Correspondingly, phenotype labels within manifestation $y$ can be classified as follows:
\begin{equation*}
    y_i = \begin{cases}
         0 & \text{if } \text{ph}_{i,\alpha},\\
         1 & \text{if } \text{ph}_{i,\beta} \text{ or } \text{ph}_{i,\gamma} \text{ or } \text{ph}_{i,\delta},
    \end{cases}
\end{equation*}
where the labels are determined by the presence of specific phenotype-defining features.
In Section \ref{sec:syn_data_gen} we will provide specific choices for these phenotype functions in Equation \eqref{eq:phenotype}.

\subsection{Predictive-Clustering Algorithm Generalizable to Phenotypes}
\label{simplepipeline}

\begin{figure*}[htbp]
\floatconts
  {modeloverview}
  {\caption{Predictive-Clustering Algorithm Generalizable to Phenotypes.}}
  {\includegraphics[width=.9\linewidth]{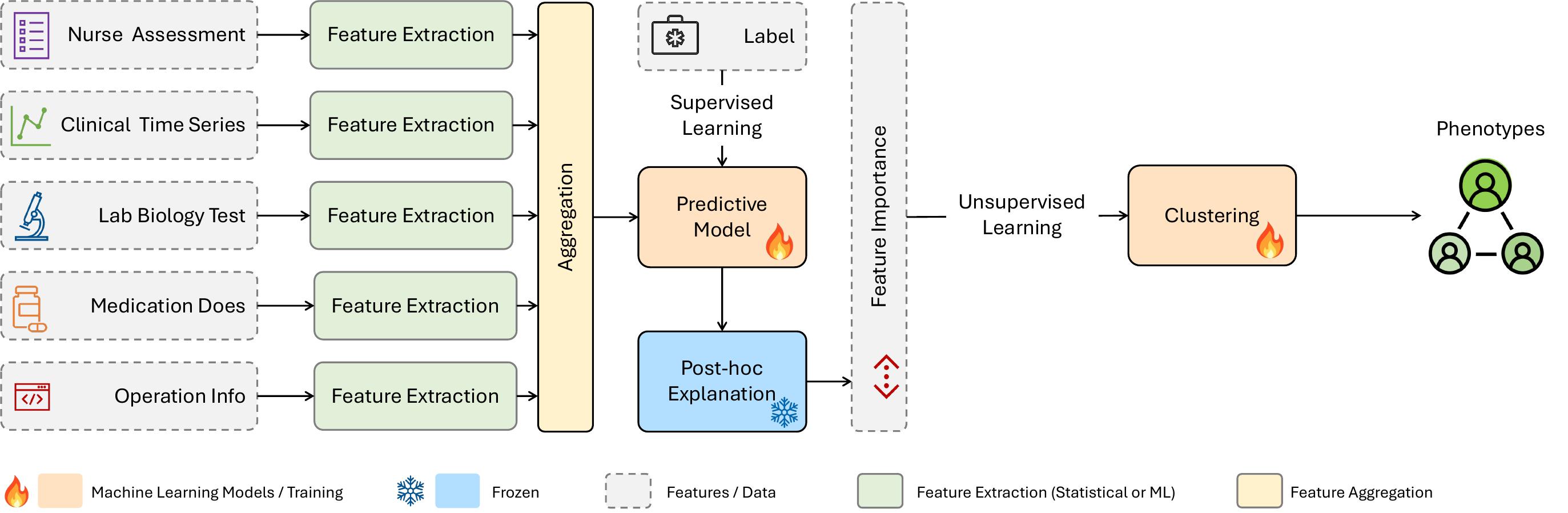}}
\end{figure*}

To test our hypothesis, we develop a simple algorithm that can be generalized to any clinical phenotype identification process. The algorithm consists of three main steps: training a predictive model, performing post-hoc analysis using SHAP, and conducting phenotype clustering based on SHAP values.
\noindent\textbf{Training a predictive model:} By utilizing only the binary labels $y_i$, we train a prediction model to estimate the probability of a patient having the disease. Our goal is to learn the conditional expectation
\begin{align*}
\mu(\bs{x})=\mathbb{E}[y_i \vert \bs{x}_i= \bs{x}],
\end{align*}
where $\bs{x}_i$ represents the feature set for patient $i$.

\noindent\textbf{Compute Personalized Explanations:} After training the prediction model, we perform a post-hoc analysis to determine the level of importance of all characteristics to get personalized explanations. We employ the SHapley Additive exPlanations (SHAP) algorithm for this purpose. SHAP is a game-theoretic approach that assigns each feature an importance value, known as the SHAP value, which represents the feature's contribution to the model's prediction. The SHAP values provide a unified measure of feature importance across different models and can be used to interpret the model behavior. Please note that additional explainability techniques, like integrated gradients~\citep{integratedsundararajan17a}, can also be incorporated into our methods.

\noindent\textbf{Phenotype clustering based on SHAP Explanations:} Using the personalized explanations (SHAP values) obtained from the post-hoc analysis, we perform phenotype clustering. 
By clustering patients based on their personalized explanations, we aim to identify distinct phenotypes \textit{within} the cohort. The clustering algorithm groups patients with similar SHAP value patterns, indicating that they share common important features that contribute to their phenotype. This step allows us to find different explanations in the development of POD and to uncover potential subtypes or phenotypes within the disease cohort, providing a more granular understanding of the disease heterogeneity.

\quad \\ \noindent By combining predictive modeling, post-hoc analysis, and phenotype clustering, our algorithm offers a generalizable and actionable approach to identifying clinical phenotypes. This methodology can be applied to various clinical diseases and can aid in the discovery of meaningful patient subgroups, leading to more targeted and personalized treatment strategies.

\section{A Case Study: Peri-Operative Delirium}
\label{casestudy}


\subsection{Multi-Modal Data}

This study utilizes a dataset consisting of multi-modal and temporal electronic health record (EHR) data (Fig.~\ref{pipeline}a-b) from patients admitted to the Intensive Care Unit (ICU) of the local University Hospital between the year of 2017 and 2022. The data covers pre-, intra-, and postoperative stages, offering a comprehensive view of each patient's health journey. The dataset includes vital signs, laboratory test results, medication history, demographic details, and operation-related information.
Patients aged 65 years or older were included in the study and classified as cases (those who developed and received treatment for delirium during their ICU stay) or controls (those who did not meet the delirium criteria). To ensure data quality and ethical compliance, patients who objected to the use of their personal health data for research or had an ICU stay shorter than 24 hours were excluded. In this study, we utilize the Intensive Care Delirium Screening Checklist (ICDSC)~\citep{icdsc} as the primary measure for identifying POD. The ICDSC is among the most developed and validated tools for this purpose, offering a comprehensive framework to systematically assess and diagnose delirium in a clinical setting~\citep{gusmao2012confusion}. We define a patient as being in delirium if the ICDSC score exceeds 3 at any point during a week-long ICU stay. A detailed description of the data and their processing methodologies is provided in the Appendix~\ref{appendix_data}.

\begin{figure*}[htbp]
\floatconts
  {pipeline}
  {\caption{Comprehensive workflow for analyzing the development of 
  peri-operative delirium based on multi-modal 
  disease trajectories using explainable ML for data-driven phenotype clustering.}}
  {\includegraphics[width=.9\linewidth]{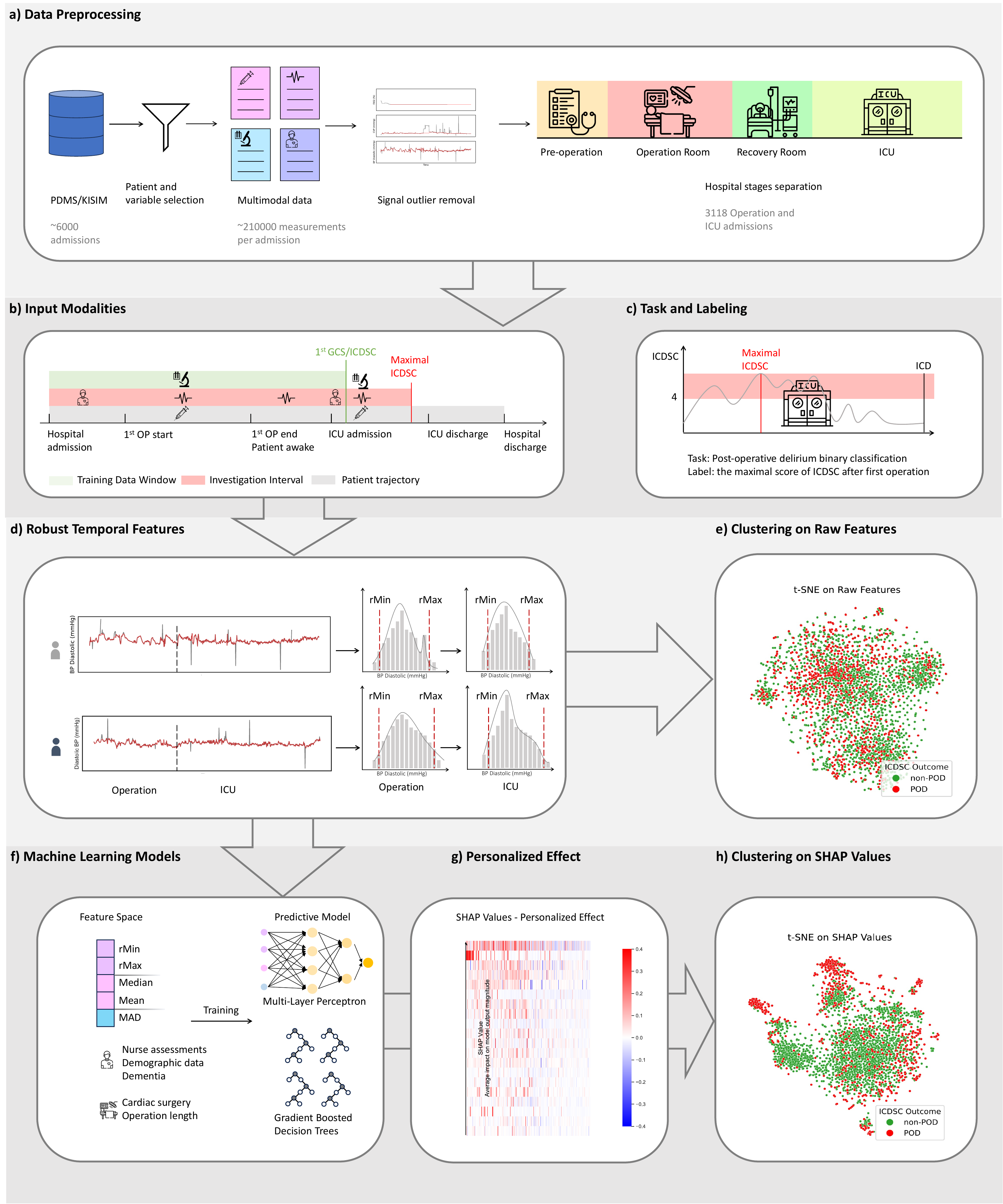}}
\end{figure*}

\subsection{Notation}

We consider longitudinal data $\mathcal{D}_N = \{\bs{X}_i, \bs{Y}_i\}_{i=1}^N$ with input $\bs{X}_i$ and outcome $\bs{Y}_i$ time series, respectively, from $N$ patients. The temporal input data involve the multivariate time series $\bs{X}_i = [\bs{x}_{i,t}]_{t=1}^T \in \mathcal{X} =\RR{T\times D}$, where $\bs{x}_{i,t} = [x_{i,t}^d]_{d=1}^D \in \RR{D}$. These input data consist of complex and temporal multi-modal data from different sources, such as demographic information, clinical nurse assessments, operation details, biometric monitoring signals, laboratory test results, medication dosages, and blood gas analyses. Further, we consider a binary outcome time series $\bs{Y}_i = [\bs{y}_{i,s}]_{s=T}^S \in \mathcal{Y}=\{0,1\}^{S-T}$. These labels correspond to a few temporal assessments indicating the physiological state of the patient such as ICDSC~\citep{icdsc} for ICU delirium. We focus on the setting, where the observed times of the outcomes are non-overlapping with the input time series, that is, we have $s \geq T$. We want to emphasize that this is a particularly challenging setting as we only observe labels far in the future. Following the data processing pipeline in Fig.~\ref{pipeline}, we introduce some specific time points to split the input time series into clinically meaningful periods, such as $\texttt{pre}$-OP, $\texttt{intra}$-OP, and $\texttt{post}$-OP. For instance, we have $\bs{\tau}_{\texttt{pre}}=[\tau_{\texttt{pre}}^0,\tau_{\texttt{pre}}^1]$ indicating the starting $\tau_{\texttt{pre}}^0$ and ending $\tau_{\texttt{pre}}^1$ time of the preoperative period, respectively. Consequently, we refer to $\bs{X}_{i,\texttt{pre}} = [\bs{x}_{i,t}]_{t=\tau_{\texttt{pre}}^0}^{\tau_{\texttt{pre}}^1}$ for the preoperative input time series. Similarly, we define $\bs{\tau}_{\texttt{intra}}$ and $\bs{\tau}_{\texttt{post}}$, leading to the intraoperative $\bs{X}_{i,\texttt{intra}}$, and postoperative time series $\bs{X}_{i,\texttt{post}}$ of patient $i$. Moreover, we introduce the corresponding \textit{cumulative} input time series $\bs{X}_{i,\texttt{intra}^+} = [\bs{x}_{i,t}]_{t=0}^{\tau_{\texttt{intra}}^1}$, which always start from $t=0$ and include time points up to the ending time point of the particular period ${\tau_{\texttt{intra}}^1}$. For the sake of simplicity, we omit the explicit dependency on $i$ when it is clear from the context, and use for instance $\bs{X}_i =\bs{X}$ and $\bs{Y}_i =\bs{Y}$, respectively.


\subsection{Our Prediction Goal}
\label{delir_pred_algo}
We want to estimate the probability of the temporal outcomes $Y_s$ at any time point $k < s \leq S$ 
given time-varying  inputs $\bs{X}_{t:k}$ from the time period  $t$ up to $k$. In particular, we want to learn the conditional expectation
\begin{align*}
\mu_{t:k}^s(\bs{x})=
    \mathbb{E}[
    Y_s \vert \bs{X}_{t:k}
    = \bs{x}
    ],
\end{align*}
where $\mu_{t:k}^s\colon 
\RR{(k-t)\times D}
\rightarrow \{0,1\}$
maps the different temporal inputs to the binary outcome.
For instance, we consider the stage-wise independent predictions
\begin{align*}
\mu_{\texttt{intra}}^s(\bs{x})
 =
    \mathbb{E}[
     Y_s \vert    \bs{X}_{\texttt{intra}}
    = \bs{x}
    ]
\end{align*}
with $k=\tau_{\texttt{pre}}^1$ and $t=\tau_{\texttt{pre}}^1$.
Further, for the cumulative predictions, we aim to compute
\begin{align*}
\mu_{\texttt{intra}^+}^s(\bs{x})
 =
    \mathbb{E}[
     Y_s \vert    \bs{X}_{\texttt{intra}^+}
    = \bs{x}
    ],
\end{align*} 
with $k=0$ and $t=\tau_{\texttt{pre}}^1$.
Similarly, for the pre- and postoperative time periods, the goal is to compute
$\mu_{\texttt{pre}}^s(\bs{x})$, $\mu_{\texttt{post}^+}^s(\bs{x})$, $\mu_{\texttt{post}}^s(\bs{x})$, and $\mu_{\texttt{post}^+}^s(\bs{x})$.



\subsection{Abstracting Temporal Complexity in Real-World Clinical Time Series}
Since it is challenging to learn
the full temporal distributions 
$\mu_{t:k}^{k:S}(\bs{x})=
    \mathbb{E}[
    \bs{Y}_{k:S} \vert \bs{X}_{t:k}
    = \bs{x}
    ]$
from noisy real-world longitudinal patient data~\citep{maibach2021synchronize} with many missing values,  weak signals, and complex temporal modalities, we follow a pragmatic approach by projecting the input and outcome time series by collapsing the temporal ordering with  specified  mappings $\psi\colon \{0,1\}^{S-k}\rightarrow \{0,1\}$ and $\phi\colon \RR{(k-t) \times D}\rightarrow \RR{M\times D}$, with $M \ll (k-t)$. In particular, we consider a simplified conditional expectation
\begin{align}
\label{eq:simple_appraoch}
\mu_{t:k}(\bs{x})=
    \mathbb{E}[
    \psi\left( \bs{Y}_{T:S} \right) \vert \phi\left( \bs{X}_{t:k} \right)
    = \bs{x}
    ],
    \end{align}
with $\mu_{t:k}\colon \RR{M\times D} \rightarrow\{0,1\}$. Besides practical reasons such as robustness against overfitting and noise in the data, this can be further justified in our context since during a certain time period, e.g.\ during an operation, the specific time of a certain event is often not relevant, instead, the overall distribution over time is more important. 

Recognizing that raw time series can be less explainable due to their intricate fluctuations and volume of data points, we turn to distribution abstraction guided by clinical knowledge to extract information from time series data. It serves as an approach to distilling these complexities into a more comprehensible format. This technique simplifies the vast array of information contained in time series by summarizing it into statistical distributions, which encapsulate the essential characteristics of the data over specific time intervals and enhance explainability. Furthermore, extracting features from the temporal distribution is particularly advantageous for harnessing the full potential of diverse data modalities in clinical EHR datasets.

For example, the overall fraction that specific variables are in a certain range, i.e. above a certain value is more important than the exact values during that time. For instance, we define hyperoxemia as an SpO$_2$ level exceeding 98\%, and when the SpO$_2$ level reaches 100\%, it is considered severe hyperoxemia.
%
%
In particular, we use different quantities to describe the temporal input distributions, that is, we use 

$$\phi\left( \bs{X}_{t:k}\right) = [\bs{x}_{\texttt{rMin}}, \bs{x}_{\texttt{rMax}}, \bs{x}_{\texttt{Median}}, \bs{x}_{\texttt{Mean}}, \bs{x}_{\texttt{MAD}}]$$
for the inputs, where $\texttt{rMin}$ and  $\texttt{rMax}$  correspond to the $5\%$ and $95\%$ quantiles, respectively, and $\texttt{MAD}$ to the \textit{median absolute deviation}. For the outcome, we use the maximum of the observed labels, that is,

$$\psi\left( \bs{Y}_{T:S}\right) = \max(\bs{Y}_{T:S}),$$
describing the most severe event in the postoperative time span. Therefore, we aim to learn the three stage-wise independent conditional expectations $\mu_{\texttt{pre}}(\bs{x})$, $\mu_{\texttt{intra}}(\bs{x})$, and $\mu_{\texttt{post}}(\bs{x})$, as well as the cumulative $\mu_{\texttt{pre}^+}(\bs{x})$, $\mu_{\texttt{intra}^+}(\bs{x})$, and $\mu_{\texttt{post}^+}(\bs{x})$. Those are obtained by plugging-in the corresponding inputs, for instance we have
$\mu_{\texttt{intra}}(\bs{x})=
    \mathbb{E}[
    \psi\left( \bs{Y}_{T:S} \right) \vert \phi\left( \bs{X}_{\texttt{intra}} \right)
    = \bs{x}
    ]$
    for $\bs{X}_{\texttt{intra}}$.


\subsection{Prediction Models}
\label{model}
For learning the conditional expectation in \eqref{eq:simple_appraoch}
    $\mu_{k:t}(\bs{x})\colon \RR{M\times D} \rightarrow \{0,1\}$,
we use different ML classification models to get an estimate
\begin{align}
\label{eq:ML_estimate}
\hat{\mu}_{t:k}(\bs{x})\colon 
\RR{M\times D}
\rightarrow \{0,1\}.
\end{align}
In particular, we consider the three stage-wise independent models
$\hat{\mu}_{\texttt{pre}}(\bs{x})$, $\hat{\mu}_{\texttt{intra}}(\bs{x})$, $\hat{\mu}_{\texttt{post}}(\bs{x})$
as well as the cumulative 
$\hat{\mu}_{\texttt{pre}^+}(\bs{x})$, $\hat{\mu}_{\texttt{intra}^+}(\bs{x})$, $\hat{\mu}_{\texttt{post}^+}(\bs{x})$. We train five machine learning models for each prediction task: logistic regression, multilayer perception, random forest, gradient boosting, and extreme gradient boosting. Logistic Regression is ideal for binary classification tasks, particularly in medical outcome prediction~\citep{song2023comparison}. It models the probability of a binary response based on one or more predictor variables through a logistic function. The Multilayer Perceptron (MLP)~\citep{popescu2009multilayer} is a class of feedforward neural network that consists of at least three layers of nodes: an input layer, one or more hidden layers, and an output layer. MLP is trained in a supervised learning setup~\citep{rumelhart1986learning}, making it capable of learning complex non-linear models and patterns in the data, making it ideal for regression and classification tasks. Random Forest~\citep{breiman2001random} is an ensemble learning method that constructs a multitude of decision trees at training time and output the mode of the classes (classification) or mean prediction (regression) of the individual trees. It is capable of running efficiently on large datasets and handling thousands of input variables. Gradient Boosting~\citep{friedman2001greedy} is a machine learning technique that builds models in a stage-wise fashion while allowing the optimization of an arbitrary differentiable loss function. It constructs new models that predict the residuals or errors of prior models and then combines them to make the final prediction, effectively improving accuracy with each iteration. This method is versatile and powerful, capable of tackling both regression and classification tasks with high efficiency. XGBoost (eXtreme Gradient Boosting)~\citep{chen2016xgboost} is an advanced implementation of gradient boosting algorithms, designed for speed and performance. It is highly efficient, scalable, and portable, offering solutions for both regression and classification problems by constructing an ensemble of decision trees sequentially, focusing on optimizing prediction accuracy and handling missing data.

\subsection{Personalized Explainability}\label{sec:personalized_shap}
The learned
ML classification models
$\hat{\mu}_{t:k}(\bs{x})\colon 
\RR{M\times D}
\rightarrow \{0,1\}$
in 
\eqref{eq:ML_estimate}
can be used
for 
analysing the 
influence of the input features
$\bs{x}\in \RR{M\times D}$.
In particular,
for a given trained ML model, we can compute the 
SHAP values
for each input feature
$x^{m,d}$, that is,
\begin{align*}
\label{eq1}
    \theta^{m,d}(\bs{x})
    =
    \mathbb{E}
    \left[
    \hat{\mu}(\bs{X}) \vert
X^{m,d} = x^{m,d}
    \right]
    -
     \mathbb{E}
    \left[
    \hat{\mu}(\bs{X}) 
    \right],
\end{align*}
indicating the difference between the expected baseline 
$\mathbb{E}
    \left[
    \hat{\mu}(\bs{X}) \right]$ and the expected outcome 
  $  \mathbb{E}
    \left[
    \hat{\mu}(\bs{X}) \vert
X^{m,d} = x^{m,d}
    \right]$,
    when changing feature $X^{m,d}=x^{m,d}$, for which we refer to~\cite{shap1,shap3,shap2}.
In particular, we 
%
consider the \textit{personalized} SHAP value
$$\theta^{m,d}_i(\bs{x}_i)
\colon \RR{M\times D} 
\rightarrow
\RR{}$$
when using 
the inputs
$\bs{x}_i\in \RR{M\times D}$ of patient $i$. Therefore, we can define the complete personalized explanation
$\bs{\theta}_i = [\theta_i^{m,d}]_{m=M,d=D}
\in \RR{MD}$
and further
all
personalized values of a cohort
$\bs{\Theta}
=
\left[
\bs{\theta}_i\right]_{i=N}\in \RR{MD\times N}$
as illustrated in Figure
  \ref{overal_results}.
  Note that we can compute personalized explanations for different
$\bs{\Theta}_{t:k}$ corresponding to the trained ML classifier 
$\hat{\mu}_{t:k}(\bs{x})$ in the time-period between $t$ and $k$.


\subsection{Clustering of Explainability Space}

We can use the personalized SHAP (i.e. feature importance)
$\theta^{m,d}_i$
summarized in
$\bs{\Theta}\in \RR{MD\times N}$
to 
answer data-driven question 
whether there are different 
phenotypes
\textit{explaining} the development of delirium.
In particular, we aim to find $K$ subgroups in an unsupervised manner based on the personalized 
explanations
which are important in distinguishing whether or not the patient develops postoperative delirium.
This enables us to
train a clustering algorithm on the personalized explanations $
\left[
\bs{\theta}_i\right]_{i=N_{train}}\in \RR{MD\times N_{train}}$
in the training set yielding a  clustering function
\begin{align*}
    \Pi(\bs{\theta}) \colon \RR{MD} \rightarrow \{1,\ldots,K\}
\end{align*}
for a patient's explanation
$\bs{\theta}\in \RR{MD}$.
More specifically, we can define the temporal clusters
$    \Pi_{t:k}(\bs{\theta}_{t:k})$
based on the personalized explanations $\bs{\Theta}_{t:k}$ corresponding to the trained ML classifier $\hat{\mu}_{t:k}(\bs{x})$. Note that this clustering algorithm is rather different from a clustering algorithm trained on raw inputs $\bs{x}\in \RR{M\times D}$ or raw temporal inputs $\bs{x}\in \RR{(k-t)\times D}$. More importantly, clustering using personalized SHAP scores is more robust and less prone to overfitting, and second, it finds different clusters or subgroups in the explanations of the learned mapping of inputs to delirium labels, allowing to draw data-driven hypotheses for unsupervised phenotypes. By applying hierarchical clustering~\citep{murtagh2014ward} to the personalized SHAP scores, our proposed method is capable of finding meaningful phenotypes with both synthetic and real-world delirium data, as demonstrated in the later Sec.~\ref{results}.

\section{Experiments and Results}
\label{results}

\subsection{Synthetic Data}

\subsubsection{Synthetic Data Generation}
\label{sec:syn_data_gen}

In this part of the study, we focus on the generation of a synthetic dataset involving
$\bs{X} \in \mathbb{R}^{N \times D}$ and $\bs{y}\in \RR{N}$, where $N$ denotes the number of samples and $D = D^{\texttt{shared}}+ D^{\texttt{informative}}+ D^{\texttt{noisy}}$ represents the number of features, as outlined in Section \ref{hypothesis}.
We assume that the $D$ features can represent any kind of clinical data, including clinical time series, EHR data, or multi-omics data. 
The main purpose of the generation of the synthetic data is the possibility to compare the ground truth phenotype and the corresponding influential features with the estimated, as this can never be evaluated on real-world data. 

Within the input feature space, we identify a critical subset of $D^{\texttt{informative}}$
for each phenotype.
We define a simple structural equation model (SEM)~\citep{sem} so that the informative features significantly influence the 
phenotype labels $\text{ph}_{i,z}\in \{0, 1\} $ in Equation \eqref{eq:phenotype} for all phenotypes $z$ and patients $i$. To mimic delirium phenotypes in our synthetically generated data, we employ the predefined set of informative features as the basis for generating them where $y_i=1$, as previously hypothesized in Sec.~\ref{hypothesis}. Each sample is assigned to a phenotype based on its randomly generated feature profile. This design approach is driven by the complex, multi-causal aspects of postoperative delirium, focusing on identifying predictive clinical indicators within a huge dataset. 

In the synthetic experiment, we define a setting with $N=3000$ samples and $D=30$ features, sampling each feature $x_d\sim \mathcal{N}(0,1)$ from a standard Gaussian distribution, where $x_d$ represents the $d^{\text{th}}$ feature. We introduce one predefined phenotype $\alpha$ as the negative samples in the synthetic cohort and three predefined phenotypes $\{\beta, \gamma, \delta\}$ as the positive samples of the cohort. For the sake of simplicity,  we set the number of informative features $D^{\text{informative}} = 2$ and the number of shared features $d^{\text{shared}} = 1$. The cohort assignment criteria are outlined in the Algorithm~\ref{alg:cohort_assignment}.

In this experiment, we first employ a gradient boosting classifier as the prediction algorithm, trained on ${\bs{X}, y}$. After this classification, we compute the SHapley Additive exPlanations (SHAP) values across the full cohorts to uncover the feature contributions towards the predictive outcomes. To visualize the results, we leverage the t-distributed Stochastic Neighbor Embedding (t-SNE) algorithm~\citep{tsne}. This technique allows us to project the high-dimensional raw feature space and feature-importance space into a lower-dimensional space, facilitating the visualization of data clustering and the relationships between samples. Furthering our analysis, we apply clustering algorithms to the feature-importance space, varying the number of clusters to explore the emergence subgroups within our synthetic dataset. This exploration aims to simulate the process of identifying latent phenotypes or subpopulations in clinical data, where the underlying subgroup structures are not directly observed. By adjusting the cluster count, we can observe the development of phenotypes. To evaluate their clustering effectiveness, we define the correct rate (\textbf{cr}) as dividing the number of samples that are accurately categorized according to their actual phenotype by the overall number of samples within the cluster, following the approach described in~\citep{subgroup4}. This rate measures the accuracy with which samples are assigned to their true categories within a given cluster.


{\centering
\begin{minipage}{.65\linewidth}
\begin{algorithm}[H]
\caption{Phenotype Assignment Criteria in a Python-like Style}
\label{alg:cohort_assignment}
\definecolor{codeblue}{rgb}{0.25,0.5,0.5}
\lstset{
  backgroundcolor=\color{white},
  basicstyle=\fontsize{7.2pt}{7.2pt}\ttfamily\selectfont,
  columns=fullflexible,
  breaklines=true,
  captionpos=b,
  commentstyle=\fontsize{7.2pt}{7.2pt}\color{codeblue},
  keywordstyle=\fontsize{7.2pt}{7.2pt},
}
\begin{lstlisting}[language=python]
import numpy as np

def f_alpha(x1, x2, x3):
    # Conditions for f_alpha phenotype
    return np.logical_and(x1 < 0, np.logical_and(x2 < 0, x3 < 0))

def f_beta(x10, x11, x12):
    # Conditions for f_beta phenotype
    return np.logical_and(np.logical_or(x10 > 0.5, x11 > 0.5), x12 > 0.5)

def f_gamma(x10, x13, x14):
    # Conditions for f_gamma phenotype
    return np.logical_and(x10 <= 0.5, np.logical_and(x13 > 0.5, x14 <= 0.5))

def f_delta(x10, x15, x16):
    # Conditions for f_delta phenotype
    return np.logical_and(x10 <= 0.5, np.logical_and(x15 <= 0.5, x16 > 0.5))
\end{lstlisting}
\end{algorithm}
\end{minipage}
\par
}

\subsubsection{Results and Discussion}

\paragraph{Analysis of clustering challenges in raw feature space.}
As a comparison, we first try to cluster the raw features of the synthetic data to identify phenotypes. However, as shown in Fig.~\ref{syndata1}, the clustering algorithm was unable to effectively distinguish between different phenotypes in the raw feature space. This suggests that the raw features may contain a lot of noise, making it challenging for the clustering algorithm to directly discover phenotype characteristics.

\begin{figure*}[htbp] 
\floatconts
  {syndata1}
  {\caption{T-SNE visualization of (a) raw features, (b) SHAP values and the ground truth outcomes for POD, and (c) SHAP~\citep{shap1} values with ground truth phenotype labels, (d)-(f) phenotype clustering with different number of clusters.}}
  {\includegraphics[trim={0cm 20cm 0cm 0cm},width=0.6\linewidth]{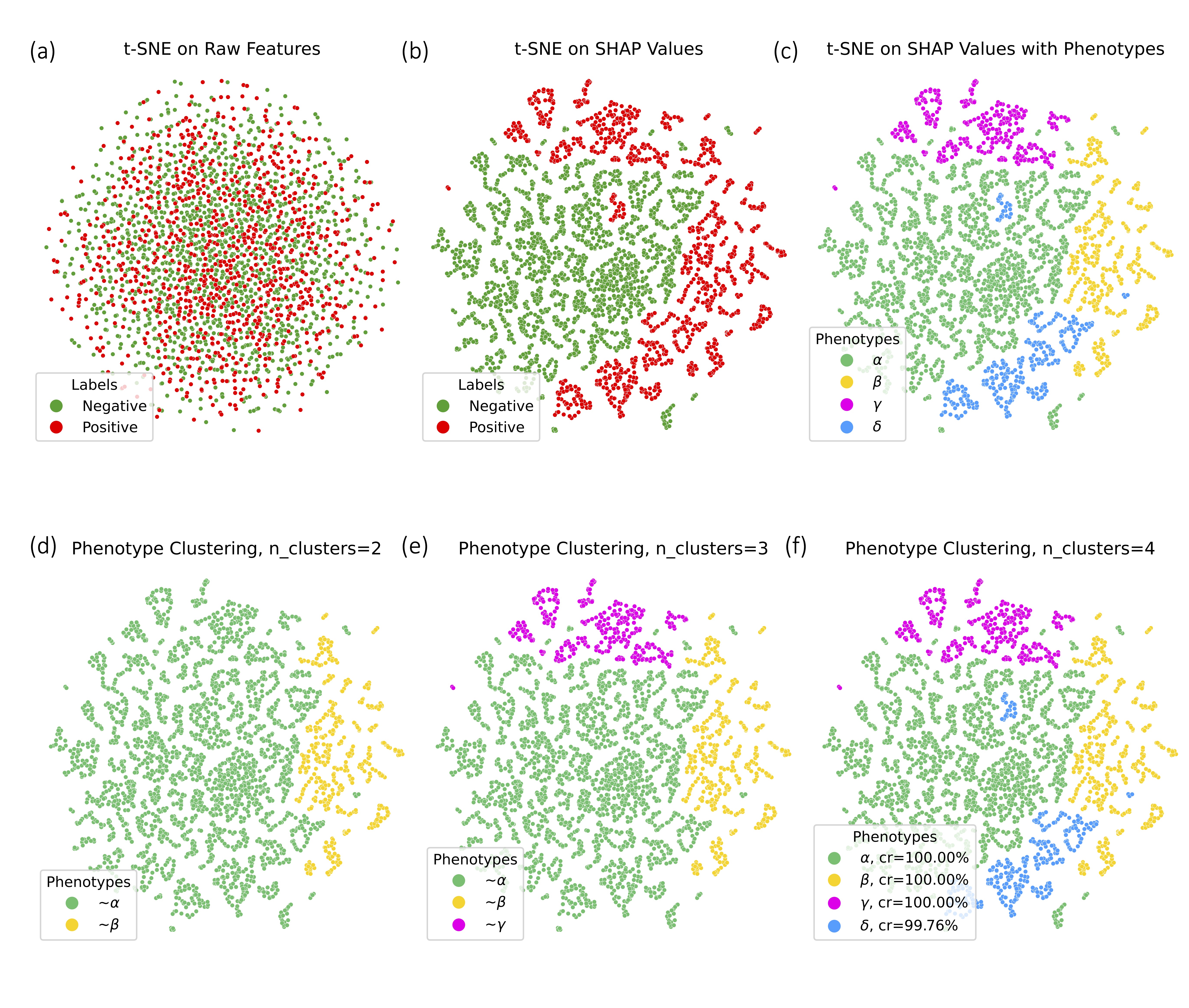}}
\end{figure*}

\paragraph{Predictive model training and results.}
To address the limitations of clustering in the raw feature space, we trained a classifier using the synthetic data. The classifier achieved an AUROC of 99\%, indicating its effectiveness in classifying different patient cohorts based on our synthetic data.

\paragraph{Phenotype clustering outcomes.}
To further investigate the phenotype characteristics, we computed SHAP values on the trained classifier and applied hierarchical clustering~\citep{murtagh2014ward} on the SHAP values. As illustrated in Fig.~\ref{syndata1}, the clustering algorithm was able to gradually identify phenotypes within the feature-importance space. This suggests that the SHAP values capture the important features and their contributions to the classifier's predictions, enabling the discovery of phenotype-specific patterns.

\paragraph{SHAP analysis to prove our hypothesis}
We hypothesized that the most important features for different phenotypes vary and consist of both informative features specific to each phenotype and shared ones common across all of them. To validate this hypothesis, we performed a SHAP analysis on the trained classifier. The results, as shown in Fig.~\ref{syndata2}, demonstrate that the top important features indeed differ for each phenotype, with a combination of phenotype-specific informative and shared features. This finding supports our hypothesis and highlights the importance of considering both unique and common features when characterizing phenotypes.

In summary, our analysis reveals that clustering in the raw feature space is challenging due to potential noise and the complexity of phenotype characteristics. By training a predictive model and leveraging SHAP values, we were able to effectively identify phenotypes and validate our hypothesis regarding the importance of phenotype-specific and shared features. These findings provide valuable insights into the underlying patterns and characteristics of different phenotypes in our synthetic data. We want to emphasize again that comparing the ground-truth phenotypes and true biomarkers can only be evaluated in these synthetic data and never in the real-world data.

\begin{figure*}[ht!] 
\floatconts
  {syndata2}
  {\caption{SHAP~\citep{shap1} analysis for different phenotypes within the cohorts where $y_i=1$.}}
  {\includegraphics[trim={0cm 1cm 0cm 0cm}, width=\linewidth]{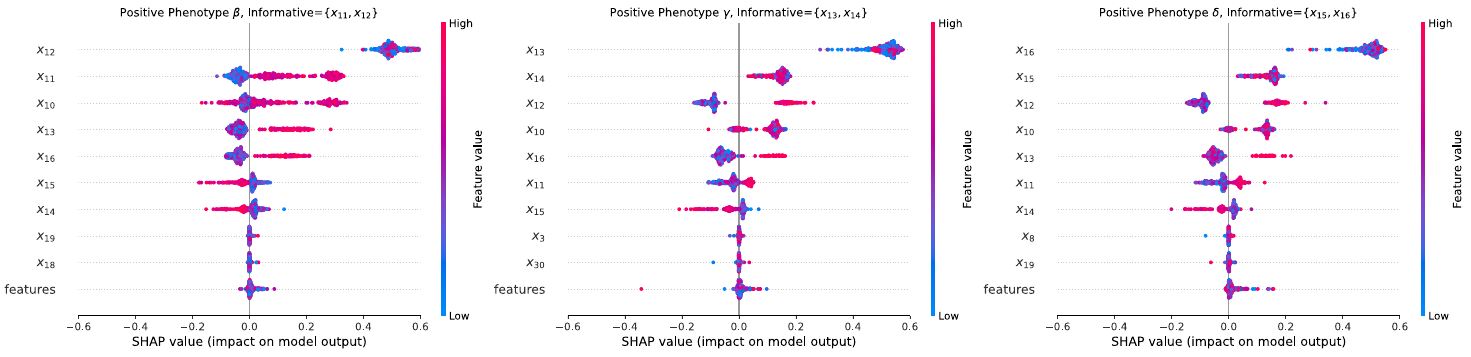}}
\end{figure*}

\subsection{Delirium Data}

\subsubsection{Experimental Settings}

Based on the patient's journey through the hospital, we have divided their trajectory into three independent stages, illustrated in Fig.~\ref{pipeline}(a)-(b). The pre-op stage $\bs{\tau}_{\texttt{pre}}$ contains the time from hospital admission until the beginning of the first operation. The intra-op stage $\bs{\tau}_{\texttt{intra}}$ refers to the duration of the first operation itself. Lastly, the post-op stage $\bs{\tau}_{\texttt{post}}$ covers the period following the first operation up until the seventh day of the patient's stay in the Intensive Care Unit (ICU). Additionally, we introduce cumulative stages—pre$^+$-OP, intra$^+$-OP, and post$^+$-OP—which incorporate data from each preceding stage, respectively, allowing for a progressively comprehensive analysis of the patient's condition and risks at each step of their hospital stay. At each cumulative hospital stage, we utilize a comprehensive feature set, including static patient demographics, operation specifics, multi-scale summaries of time series distributions, and various clinical variables' measurement intensity.


%

For the evaluation of our prediction models in Sec.~\ref{model}, we computed the Area Under the Receiver Operating Characteristic (AUROC) and the Area Under the Precision-Recall Curve (AUPRC) as our performance metrics. The AUROC provides a comprehensive measure of the model's ability to distinguish between classes, while the AUPRC offers valuable insights into the model's precision and recall, making them ideal for evaluating the performance of our models.  The final evaluation of  the trained models is based on their average performance on the test sets across the 10 folds.










\subsubsection{Results and Discussion}

\begin{table*}[ht!]
\footnotesize
\centering
\caption{Models Performance for different hospital stages}
\resizebox{\textwidth}{!}{
\begin{tabular}{c|c|c|c|c|c|c}
\hline
Models & \multicolumn{2}{c}{pre$^+$-OP} & \multicolumn{2}{|c|}{intra$^+$-OP} & \multicolumn{2}{c}{post$^+$-OP} \\ \hline
& AUROC & AUPRC & AUROC & AUPRC & AUROC & AUPRC \\

\hline
Logistic Regression & $0.698 \pm 0.023$ & $0.454 \pm 0.022$& $0.666 \pm 0.029$ & $0.427 \pm 0.035$& $0.659 \pm 0.023$ & $0.428 \pm 0.033$ \\
Multilayer Perceptron   & $0.704 \pm 0.029$ & $0.452 \pm 0.030$& $0.669 \pm 0.030$ & $0.429 \pm 0.029$& $0.667 \pm 0.025$ & $0.433 \pm 0.035$ \\
Random Forest & $0.721 \pm 0.029$ & $0.499 \pm 0.037$& $0.712 \pm 0.034$ & $0.474 \pm 0.036$& $0.724 \pm 0.031$ & $0.490 \pm 0.037$ \\
XGBoost & $0.675 \pm 0.014$ & $0.445 \pm 0.019$& $0.690 \pm 0.024$ & $0.454 \pm 0.033$& $0.731 \pm 0.021$ & $0.512 \pm 0.028$ \\
Gradient Boosting & $0.714 \pm 0.026$ & $0.485 \pm 0.029$& $0.704 \pm 0.027$ & $0.466 \pm 0.029$& $0.743 \pm 0.019$ & $0.533 \pm 0.029$ \\

\hline
Models & \multicolumn{2}{c}{pre-OP} & \multicolumn{2}{|c|}{intra-OP} & \multicolumn{2}{c}{post-OP} \\ \hline
& AUROC & AUPRC & AUROC & AUPRC & AUROC & AUPRC \\

\hline
Logistic Regression & $0.691 \pm 0.024$ & $0.426 \pm 0.022$& $0.561 \pm 0.029$ & $0.322 \pm 0.026$& $0.592 \pm 0.018$ & $0.360 \pm 0.016$ \\
Multilayer Perceptron   & $0.696 \pm 0.029$ & $0.436 \pm 0.029$& $0.551 \pm 0.030$ & $0.318 \pm 0.026$& $0.598 \pm 0.030$ & $0.362 \pm 0.026$ \\
Random Forest & $0.719 \pm 0.029$ & $0.498 \pm 0.037$& $0.576 \pm 0.026$ & $0.326 \pm 0.026$& $0.653 \pm 0.030$ & $0.419 \pm 0.041$ \\
XGBoost & $0.680 \pm 0.019$ & $0.449 \pm 0.019$& $0.556 \pm 0.019$ & $0.316 \pm 0.015$& $0.652 \pm 0.024$ & $0.426 \pm 0.021$ \\
Gradient Boosting & $0.714 \pm 0.026$ & $0.474 \pm 0.026$& $0.558 \pm 0.019$ & $0.319 \pm 0.018$& $0.674 \pm 0.025$ & $0.462 \pm 0.029$ \\

\hline
\end{tabular}
}
\label{overalltab}
\end{table*}

\begin{figure*}[ht!]
\floatconts
  {overal_results}
  {\caption{Predictive model performance for different hospital stages and personalized explanation with cohort with \texttt{post$^+$-op} features.}}
  {\includegraphics[trim={0cm 2cm 0cm 0cm}, width=\linewidth]{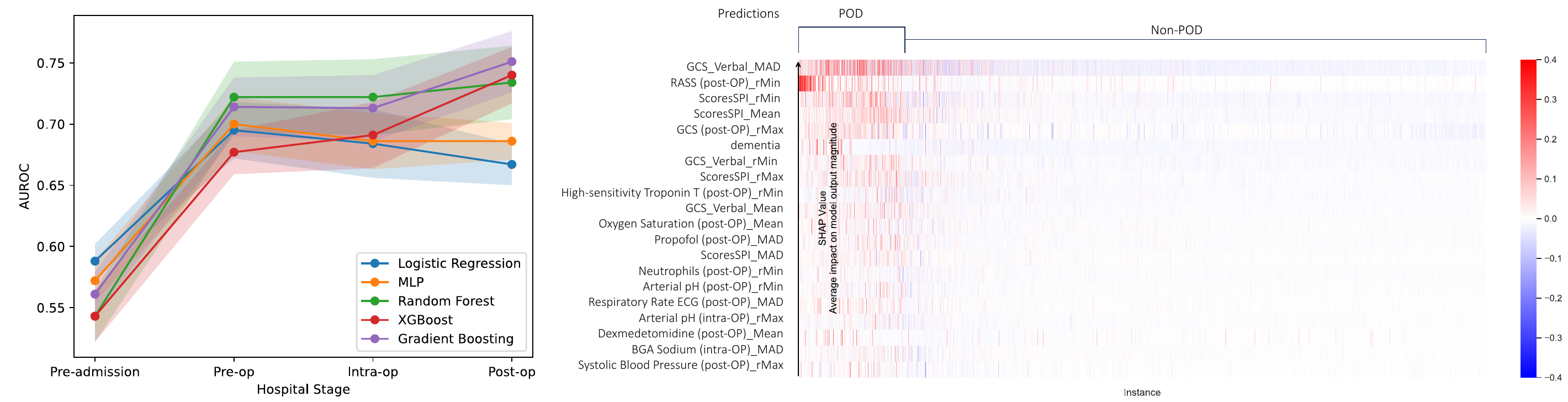}}
\end{figure*}

\paragraph{Prediction results and personalized risk factor explanations.}
Following the pipeline described in Sec.~\ref{simplepipeline} and the methods in Sec.~\ref{delir_pred_algo}, we apply different predictive machine learning models described in Sec.~\ref{model} for different hospital stages. In the comparative analysis of predictive models for POD, Gradient Boosting has the highest overall performance across all models during every hospital stage, reflecting its capacity to model diverse data modalities and stages of hospital care. The comprehensive utilization of data spanning across all hospital stages yielded optimal performance of AUROC $0.743 \pm 0.019$ and AUPRC $0.533 \pm 0.029$. When considering only  intra-operative data, a marginal decrease in predictive accuracy is observed for trained models. This could suggest the presence of noise within intra-operative patient data. Moreover, the cumulative models trained on all previous history showed consistently better performance compared to non cumulative counterparts (i.e. ones trained on start and end of each stage - see Table \ref{overalltab}).  Overall, Gradient Boosting model achieved a consistent improving performance, across all stages, showcasing its robustness in modeling multi-modal data including the noisy interval of intra-operative phase (Figure \ref{overalltab}).

    

We employ our personalized SHAP scores as described in Sec.~\ref{sec:personalized_shap} to elucidate the attribution of various features in the model prediction.
This approach of personalized explanation allows us to further explore each individual's case, providing a more detailed understanding of the diverse factors influencing their health outcomes. 
By examining the specific contributions of different features to the risk of developing delirium for each patient, we observe that the top 5 risk factors come from clinical assessments, including \texttt{GCS} (Glasgow Coma Scale)~\citep{gcs}, \texttt{RASS} (Richmond Agitation-Sedation Scale)~\citep{rass}, and \texttt{SPI} (Self-Care Index, German: Selbstpflege-Index)~\citep{spi-delirium-1}, as shown in Fig.~\ref{overal_results}. Variable risk factors among different individuals provide insights into their unique clinical conditions. Furthermore, this method helps identify different cohorts of patients with respect to the causes of delirium, enabling targeted interventions and more effective management strategies tailored to the unique risk profiles of different groups. 
\begin{figure}[ht!]
\centering
\floatconts
  {raw-feature-space}
  {\caption{T-SNE visualization of a) raw features, b) SHAP values and the ground truth outcomes for POD, and c) SHAP values with assigned phenotype clusters.}}
  {\includegraphics[trim={0cm 2cm 0cm 0cm},clip, width=0.8\linewidth]{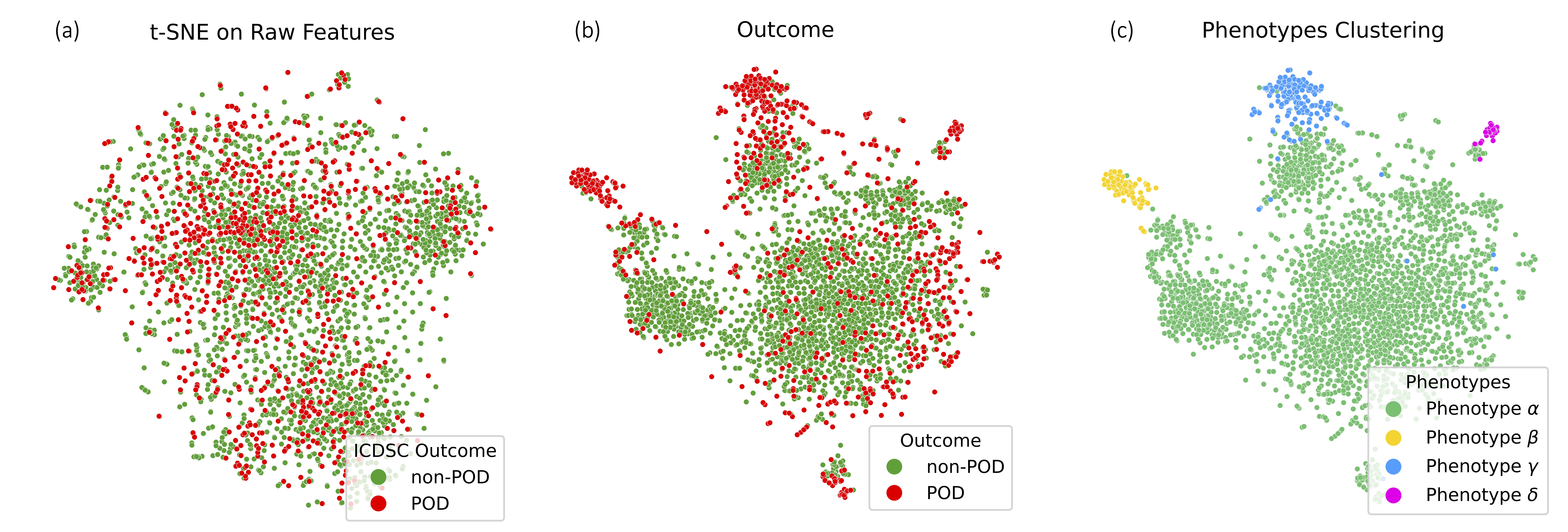}}
\end{figure}

\paragraph{Analysis of clustering in the raw feature space.} 
As a comparison, we first attempted to cluster the raw features of the real delirium data to identify the phenotypes. However, as shown in Fig.~\ref{raw-feature-space}(a) with more results with different dimension reduction algorithms in Appendix Fig.~\ref{appendix_raw}, the clustering algorithm was unable to effectively distinguish between different phenotypes in the raw feature space. The results show that there is no clear separation based on raw features alone, indicating that relying solely on raw features is insufficient to discover delirium phenotypes. This suggests that raw features may contain noise and complex interactions as we hypothesized, making it challenging for the clustering algorithm to directly uncover the characteristics of the phenotype.

\paragraph{Identifying patient subgroups through clustering in feature-importance space.}
To further investigate the phenotype characteristics within the delirium data, following the algorithm described in Sec.~\ref{simplepipeline}, we computed SHAP values on the trained Gradient Boosting classifier and applied hierarchical clustering on the SHAP values. As illustrated in Fig.~\ref{raw-feature-space}(b) and (c), the clustering algorithm was able to identify distinct patient subgroups within the feature-importance space. The t-SNE visualization of the SHAP values shows clear separations between the identified clusters, indicating that the SHAP values capture the important features and their contributions to the classifier's predictions, enabling the discovery of phenotype-specific patterns and explanations.

\paragraph{Phenotype-based Explainability analysis.}

\begin{figure*}[ht!] 
\floatconts
  {shap_delir_phenotypes}
  {\caption{SHAP analysis for different phenotypes within delirium cohorts.}}
  {\includegraphics[trim={0cm 1cm 0cm 0cm}, width=0.9\linewidth]{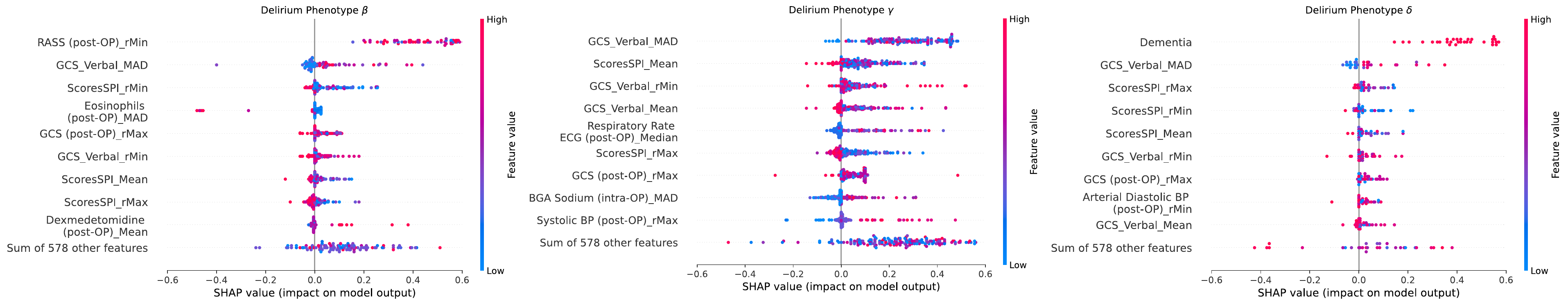}}
\end{figure*}

\begin{figure*}[ht!] 
\floatconts
  {shap_delir_phenotypes_scatter}
  {\caption{Bridging SHAP analysis and clinical raw feature interpretation}}
  {\includegraphics[trim={0cm 10cm 0cm 4cm},clip,width=\linewidth]{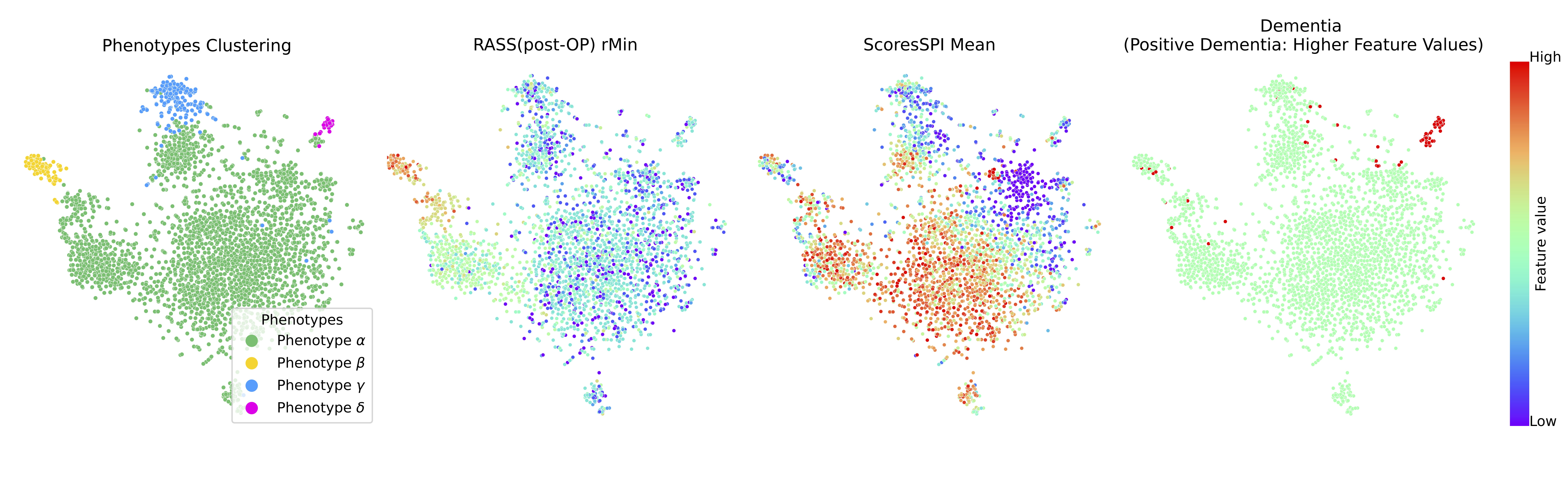}}
\end{figure*}


In the analysis of phenotype characteristics using SHAP, our study has identified distinct risk factors that play a dominant role in the manifestation of delirium across different phenotypes. This underscores the clinical diversity observed in delirium presence and suggests that each phenotype may be driven by unique underlying mechanisms or pathways~\citep{delirium_subgroup4}. For instance, as shown in Fig.~\ref{shap_delir_phenotypes}, in phenotype $\beta$, we can find that low eosinophil count~\citep{dunphy2021mixed-ec-dementia} is one of the risk factors of POD and the use of dexmedetomidine~\citep{flukiger2018dexmedetomidine} is correlated to reduced POD. The prominence of eosinophil count~\citep{eosinophil1,eosinophil2} suggests a potential link between immune system dysfunction and the development of delirium. Additionally, the identification of dexmedetomidine as a significant factor indicates that the choice of sedative agents may have a significant impact on the risk of delirium in this subgroup. Please note that there are some cases where the usage of dexmedetomidine will develop POD. We aim to investigate the multi-causal factors contributing to these cases in the future work. In phenotype $\gamma$, the respiratory rate ECG and systolic blood pressure~\citep{sbp1,sbp2} in the ICU room are more important for the development of delirium. This suggests that phenotype $\gamma$ may be more associated with cardiovascular and respiratory dysfunction, potentially indicating a greater influence of physiological stressors on the development of delirium in this subgroup. A particularly notable finding is the strong association between dementia and delirium in phenotype $\delta$. In this group, dementia was a common condition among all patients, marking it as a key risk factor. This observation is critical because the correlation between dementia and POD has long been established, as evidenced by results from the National Inpatient Sample (NIS) database, where dementia patients had a higher POD ($15.4\%$ vs $1.5\%$, $p=<0.001$) as compared with patients with no dementia~\citep{dementia1}. Therefore, this finding strongly supports the validity of our approach.


Meanwhile, our approach seeks to connect these interpretable insights back to the original raw features from which they were derived. As shown in Fig.~\ref{shap_delir_phenotypes_scatter}, based on the SHAP analysis, we identified three clinical features that influence the occurrence of delirium. A higher robust minimal value of RASS can be related to a higher incidence of delirium in phenotype $\beta$. The lower mean value of the SPI score is probably associated with a higher incidence of delirium in the phenotype $\gamma$. As expected, in the phenotype $\delta$, almost all patients have dementia. By examining the SHAP values within each cluster, we can gain insights into the unique risk factors and feature contributions that characterize each patient subgroup. This analysis reveals that patients within the same subgroup share similar risk profiles, with specific combinations of clinical assessments and biomarkers playing an important role in their predicted risk of developing postoperative delirium.


\begin{figure*}[ht!]
\centering
\floatconts
  {hiercluster}
  {\caption{Development of subgroups through different hospital stages}}
  {\includegraphics[trim={4cm 25cm 4cm 4cm},clip, width=0.7\linewidth]{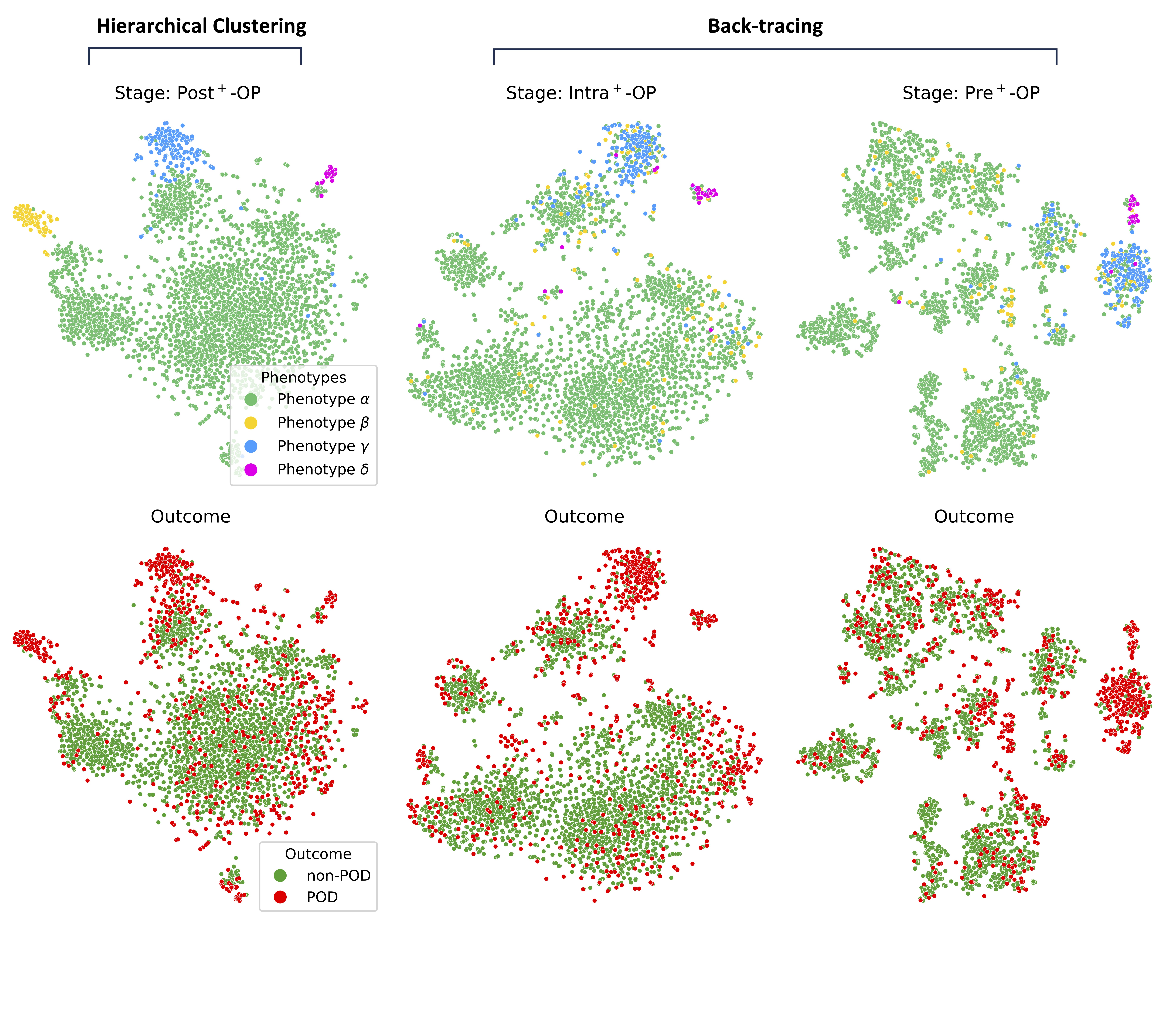}}
\end{figure*}

\paragraph{Temporal evolution of SHAP-based clusters.} Subsequently, we focused on clustering the SHAP features across different hospital stages. Fig.~\ref{hiercluster} illustrates that the \textbf{\textcolor{cluster3}{phenotype $\gamma$}} is more distinctly identifiable, indicating a unique set of risk factors that can be recognized early on. In contrast, \textbf{\textcolor{cluster2}{phenotype $\beta$}} and \textbf{\textcolor{cluster4}{phenotype $\delta$}}  initially appear dispersed within the space. However, as patients progress through the various stages of hospital care, these two subgroups gradually become more defined and separate from each other. This evolution highlights the dynamic nature of POD risk as influenced by the changing clinical landscape, demonstrating that certain risk factors become more or less prominent as the patient's condition evolves.

\section{Conclusion and Future Work}
In conclusion, our study has demonstrated that the identification of phenotypes based on feature importance values deriving from predictive models, enhances our understanding of the heterogeneity within complex clinical disease. By clustering patients in the SHAP feature importance space, we have shown that our approach outperforms clustering in the raw feature space, enabling the discovery of more clinically meaningful subtypes. We used postoperative delirium as a real-world case study to show that our methodology has broader implications for the analysis of clinical disease. Our findings highlight the potential of this approach to bridge the gap between machine learning models and their clinical application, improving the explainability and interpretability of complex clinical data. 

In our future research, we plan to create advanced models for irregular time series that are capable of dealing with the diverse types of data commonly found in healthcare environments. Additionally, we aim to focus on refining these models to achieve a more minimalistic approach that maintains high accuracy while simplifying input heterogeneity. Therefore, less but more meaningful features will further enhance the clinical usability of our models, making them more accessible to healthcare professionals and potentially extending their application to other complex neuropsychiatric conditions.

\section*{Acknowledgments}
This work is supported by the Swiss National Science Foundation (project 201184).

\newpage
\bibliography{delirium-arxiv}

\newpage
\appendix

\section{Data Description for Delirium Case Study}
\label{appendix_data}

\subsection{Raw Data and Acquisition Stage}\label{apd}

Patient data utilized in this study were systematically retrieved from the KISIM and PDMS at the local University Hospital 
between November 1, 2017, and December 31, 2022. Each patient in the dataset is assigned a unique patient ID. For different hospital visits, distinct case research IDs are allocated. A critical aspect of the dataset is the timestamp associated with each measurement, providing a chronological context to the clinical data. Meanwhile, each patient's journey throughout the hospital stay is meticulously tracked and represented through a set of unique unit codes (UnitID), distinctly identifying key locations such as the operation room, recovery room, and Intensive Care Unit (ICU).

The system categorized and coded patient statuses to accurately reflect their specific hospitalization action, encompassing categories such as:
\begin{enumerate}
    \item PreAdmission: Denotes instances in which a patient is scheduled and awaiting bed assignment.
    \item Admission: Indicating that the patient has been allocated a bed.
    \item Operation: Reflecting the periods where the patient is under anesthesia.
    \item Releasing: Signifying moments when the patient is ready for the transition, either to a different bed or care area.
    \item Discharge: Capturing the point at which the patient leaves the ICU or the operation zone.
\end{enumerate}

\subsection{Multi-Modal Data}

We leverage the rich spectrum of multi-modal and temporal electronic health record (EHR) data, encompassing pre, intra, and postoperative stages, as illustrated in Fig.~\ref{pipeline}~(a) and~(b). In particular, our method integrates diverse data types including vital signs, laboratory test results, medication history, demographic details, and specific operation-related information. By harnessing these varied data sources, our model aims to capture the rich information contained in each patient's monitoring data, providing a holistic view of their risk for developing POD.
Working with this complex data involves several steps and tasks as illustrated in Fig.~\ref{pipeline} and will be discussed in the following sections.

\subsubsection{Data Preprocessing}

\paragraph{Patient and Variable Selection:} We extracted patient data from the Patient Data Management System (PDMS) and the KISIM system in the local University hospital. Each patient's electronic health record is initially extracted using their unique patient ID. For each hospital visit, we utilize the distinct research case IDs assigned to the patients. This step ensures that each hospital visit is treated as a separate instance, accommodating patients with multiple visits. Then we extracted multimodal data from electronic health records of each hospital visit, containing demographic information, clinical nurse assessments, operation details (including type and length), and a series of intraoperative and postoperative data. This includes biometric monitoring signals, laboratory test results, medication dosages, and blood gas analyses both during the operations and in the postoperative phases in recovery rooms and ICU. The detailed selected features are reported in the appendix.

\begin{table*}[tb!]
\centering
\caption{Statistical Analysis of Total Medication Doses Administered During Surgery and in the ICU (Excluding Non-Used Medications)}
\label{tab:med}
\resizebox{\textwidth}{!}{
\begin{tabular}{l|,,.}
\toprule
Variables &\mc{POD (n=839)} & \mc{Non-POD (n=2279)} & \mc{p-value} \\
\midrule
\rowcolor{lightgray}\multicolumn{4}{l}{Intra-Operative Medication (total doses)}\\
\rowcolor{lightgray}\quad Dexmedetomidine [mg] &            0.1(0.0-0.1)\  |\  42(5.0\%) &           0.1(0.0-0.2)\  |\  156(6.8\%) &        0.503 \\
\rowcolor{lightgray}\quad Fentanyl [mcg] &    500.0(300.0-500.0)\  |\  651(77.6\%) &   500.0(300.0-500.0)\  |\  1759(77.2\%) &        0.009 \\
\rowcolor{lightgray}\quad Ketamine [mg] &         40.0(30.0-70.4)\  |\  73(8.7\%) &       50.0(30.0-80.0)\  |\  235(10.3\%) &        0.454 \\
\rowcolor{lightgray}\quad Midazolam [mg] &          3.0(2.0-5.0)\  |\  166(19.8\%) &          3.0(2.0-5.0)\  |\  395(17.3\%) &        0.712 \\
\rowcolor{lightgray}\quad Morphine Hydrochloride Hydrate &            4.0(2.0-5.0)\  |\  37(4.4\%) &            4.0(2.0-6.0)\  |\  84(3.7\%) &        0.461 \\
\rowcolor{lightgray}\quad Propofol [mg] &     140.0(80.0-546.6)\  |\  585(69.7\%) &    140.0(90.0-663.3)\  |\  1601(70.3\%) &        0.282 \\
\rowcolor{lightgray}\quad Remifentanil [mg] &  1220.0(556.2-2653.2)\  |\  118(14.1\%) &  1869.0(887.5-2770.8)\  |\  356(15.6\%) &        0.049 \\

\multicolumn{4}{l}{Post-Operative ICU Medication (total doses)}\\
\quad Dexmedetomidine [mg] &          0.7(0.2-2.0)\  |\  336(40.0\%) &           0.2(0.1-0.6)\  |\  328(14.4\%) &        <0.001 \\
\quad Fentanyl [mcg] &    200.0(100.0-350.0)\  |\  294(35.0\%) &      150.0(71.2-275.0)\  |\  530(23.3\%) &        0.718 \\
\quad Ketamine [mg] &       130.1(50.0-727.2)\  |\  48(5.7\%) &      658.7(99.1-2251.2)\  |\  134(5.9\%) &        0.046 \\
\quad Lorazepam [mg] &          2.0(1.0-6.0)\  |\  125(14.9\%) &            1.0(1.0-2.5)\  |\  123(5.4\%) &        <0.001 \\
\quad Midazolam [mg] &         5.0(2.0-15.1)\  |\  205(24.4\%) &        15.6(5.0-147.1)\  |\  286(12.5\%) &        <0.001 \\
\quad Morphine Hydrochloride [mg] &          11.0(3.0-20.5)\  |\  19(2.3\%) &            6.0(2.0-11.5)\  |\  47(2.1\%) &        0.628 \\
\quad Morphine Hydrochloride Hydrate &        13.0(6.0-24.0)\  |\  481(57.3\%) &        10.0(4.0-16.0)\  |\  1142(50.1\%) &        <0.001 \\
\quad Morphine Sulfate Pentahydrate [mg] &          35.0(32.5-37.5)\  |\  2(0.2\%) &         160.0(20.0-270.0)\  |\  5(0.2\%) &        0.382 \\
\quad Oxazepam [mg] &        127.5(15.0-146.2)\  |\  7(0.8\%) &          15.0(15.0-24.4)\  |\  12(0.5\%) &        0.011 \\
\quad Oxycodone [mg] &          10.0(5.0-21.2)\  |\  72(8.6\%) &           5.0(5.0-15.0)\  |\  176(7.7\%) &        0.490 \\
\quad Oxycodone Hydrochloride [mg] &         20.0(10.0-37.5)\  |\  75(8.9\%) &         10.0(5.0-20.0)\  |\  262(11.5\%) &        <0.001 \\

\bottomrule
\multicolumn{3}{l}{\textit{Note}: Data are presented as median (interquartile range) $|$ number of patients (\%)}

\end{tabular}}
\end{table*}
\begin{table*}[tb!]
\centering
\caption{Statistical Analysis of Average Laboratory Testing Value During Surgery and in the ICU}
\label{tab:ikc}
\resizebox{\textwidth}{!}{
\begin{tabular}{l|,,.}
\toprule
Variables &\mc{POD (n=839)} & \mc{Non-POD (n=2279)} & \mc{p-value} \\
\midrule
\rowcolor{lightgray}\multicolumn{4}{l}{Operative laboratory testing (mean)}\\
\rowcolor{lightgray}\quad Basophils       & 0.04 (0.03-0.06)\ |\  254(30.3\%) &0.04 (0.23-0.07)\ |\  482(21.1\%)&  0.42 \\
\rowcolor{lightgray}\quad Total Bilirubin       & 9.0 (5.0-14.0)\ |\  245(29.2\%)&8.0 (5.0-13.0)\ |\  453(19.9\%)&   0.73     \\
\rowcolor{lightgray}\quad C-Reactive Protein       &9.9 (1.7-58.5)\ |\  271(32.3\%)&7.6(2.0-45.0)\ |\  488(21.4\%)&  0.34 \\
\rowcolor{lightgray}\quad Eosinophils       &0.04(0.01-0.14)\ |\  254(30.3\%)&0.04(0.01-0.11)\ |\  482(21.1\%)&0.69 \\
\rowcolor{lightgray}\quad Hemoglobin       &115.0(89.8-128.0)\ |\  340(40.5\%)&112.0(89.0-130.0)\ |\  683(30.0\%)&0.97 \\
\rowcolor{lightgray}\quad Urea       &(7.4(5.5-10.9)\ |\  263(31.4\%)&7.4(5.6-10.5)\ |\  473(20.8\%)&  0.51 \\
\rowcolor{lightgray}\quad Creatinine       &89.0(67.0-131.0)\ |\  273(32.5\%)&94.0(72.0-132.3)\ |\  500(21.9\%)&0.97 \\
\rowcolor{lightgray}\quad White Blood Cells       &10.7(8.25-15.2)\ |\  340(40.5\%)&11.3(7.90-15.4)\ |\  683(30.0\%)&0.36 \\
\rowcolor{lightgray}\quad Lymphocytes       &1.04(0.62-1.60)\ |\  254(30.3\%)&1.06(0.65-1.82)\ |\  482(21.1\%)&0.30\\
\rowcolor{lightgray}\quad Magnesium       &0.80(0.72-0.90)\ |\  250(29.8\%)&0.82(0.75-0.91)\ |\  438(19.2\%)&0.08 \\
\rowcolor{lightgray}\quad Monocytes       &0.68(0.42-0.96)\ |\  254(30.3\%)&0.69(0.46-0.97)\ |\  482(21.2\%)&0.34 \\
\rowcolor{lightgray}\quad Neutrophils       &9.03(6.71-12.8)\ |\  254(30.3\%)&9.27(6.39-13.5)\ |\  482(21.2\%)&  0.40 \\
\rowcolor{lightgray}\quad Pro B-Type Natriuretic Peptide       &1699(361-7930)\ |\  92(11.0\%)&2340(754.4-5098)\ |\  155(0.07\%)& 0.35 \\
\rowcolor{lightgray}\quad Platelets       &198(148-255.5)\ |\  331(39.5\%)&189(144.9-248)\ |\  664(29.1\%)&   0.40 \\
\rowcolor{lightgray}\quad High-sensitivity Troponin T       &44(18-92.5)\ |\  256(30.5\%)&41(21-102.5)\ |\  468(20.5\%)&   0.60 \\
\multicolumn{3}{l}{Postoperative laboratory testing (mean)}\\
\quad Basophils       &     0.03 (0.02-0.05)\ |\  818(97.5\%) &0.03 (0.02-0.05)\ |\  2179(95.6\%)&  0.39 \\
\quad Total Bilirubin       & 9.0 (6.0-16.0)\ |\  564(67.2\%)&10.5 (7.0-18.0)\ |\  1123(53.7\%)&   0.67    \\
\quad C-Reactive Protein       &78.6 (40.2-138.4)\ |\  826(98.5\%)&45.6(22.0-95.1)\ |\  2228(97.8\%)&  <0.001 \\
\quad Eosinophils       &0.07(0.03-0.16)\ |\  818(97.5\%)&0.05(0.01-0.11)\ |\  2179(95.6\%)&<0.001 \\
\quad Hemoglobin       &93.5(82.2-110.5)\ |\  831(99.1\%)&103.5(87.7-118.5)\ |\  2224(98.5\%)&<0.001 \\
\quad Urea       &7.4(5.2-11.4)\ |\  793(94.5\%)&6.4(4.6-9.2)\ |\  1997(87.6\%)&  <0.001 \\
\quad Creatinine       &89.0(67.0-131.9)\ |\  827(98.6\%)&85.0(67.5-116.0)\ |\  2232(97.9\%)&<0.001 \\
\quad White Blood Cells       &11.2(8.63-14.5)\ |\  831(99.1\%)&11.1(8.77-14.3)\ |\  2244(98.5\%)&0.95 \\
\quad Lymphocytes       &1.02(0.72-1.44)\ |\  818(97.5\%)&1.02(0.72-1.44)\ |\  2179(95.6\%)&0.29\\
\quad Magnesium       &0.96(0.87-1.06)\ |\  794(94.6\%)&0.95(0.85-1.07)\ |\  2018(88.6\%)&0.46 \\
\quad Monocytes       &0.80(0.57-1.07)\ |\  818(97.5\%)&0.78(0.56-1.03)\ |\  2179(95.6\%)&0.39 \\
\quad Neutrophils       &8.91(6.78-12.0)\ |\  818(97.5\%)&9.0(6.83-11.9)\ |\  2179(96.6\%)&  0.93 \\
\quad Pro B-Type Natriuretic Peptide       &2563(807.9-6496)\ |\  401(47.8\%)&1595(538.2-5102.2)\ |\  777(34.1\%)& 0.01 \\
\quad Platelets       &169(123.8-227.2)\ |\  831(99.1\%)&170(127-223.4)\ |\  2233(97.9\%)&   0.50 \\
\quad High-sensitivity Troponin T       &108.5(36.1-603.5)\ |\  708(84.4\%)&128.5(31.0-557.7)\ |\  1818(79.8\%)&   0.36 \\
\bottomrule
\multicolumn{3}{l}{\textit{Note}: Data are presented as median (interquartile range) $|$ number of patients (\%)}
\end{tabular}
}
\end{table*}

\paragraph{Outliers Removal:}
We preprocessed the data by removing outliers or artifacts due to sensor errors, data entry mistakes, or physiological anomalies. The removal of these artifacts is performed using a combination of statistical methods, such as interquartile range method~\citep{chu2016data}, threshold-based exclusions, and manual reviews where necessary. The goal is to ensure that the data used for analysis is as accurate and clean as possible. Variables with more than $20\%$ missing values were excluded from consideration. Among considered variables, any missing values were addressed using the last observation carried forward method~\citep{locf1,locf2}. In situations where a patient had no previously recorded value for a variable, we imputed the missing data with the overall average for continuous variables.

\paragraph{Hospital Stage Separation:} We divided the patient’s hospitalization journey into three stages based on their location and the actions taking place: (1) preoperation, spanning from hospital admission to the start of the operation, (2) intraoperation, the time spent in the operation, and (3) the postoperative recovery process including the  ICU stay. We identify the patients who have had at least one operation during their hospital stay. The analysis is focused on the first operation and the immediate ICU stay that follows. 
Subsequent operations are excluded to maintain consistency in the dataset. Furthermore, a filter is applied to select patients who have an ICU stay at least one week after their first operation.

\subsubsection{Input Modalities and Study Period}

Our methods incorporate multiple data modalities corresponding to different hospital stages when training our machine learning model for delirium prediction. In the preoperation stage, data consists of detailed nurse assessments, with specifics provided in the appendix. For the operation and ICU stages, the dataset includes biometric monitoring signals, laboratory test results, medication dosages, and blood gas analyses. Additionally, the ICU stage data include nurse assessments. The model utilizes data collected before the first Glasgow Coma Scale (GCS) and Intensive Care Delirium Screening Checklist (ICDSC) assessments. This approach aims to enable clinicians to predict the potential onset of delirium before these initial assessments are conducted, facilitating early detection and prevention of delirium.



\subsubsection{Task and Labelling}

In this study, we utilize the Intensive Care Delirium Screening Checklist (ICDSC)~\citep{icdsc} as the primary measure for identifying POD, as illustrate in Fig.~\ref{pipeline}(c). The ICDSC is among the most developed and validated tools for this purpose, offering a comprehensive framework to systematically assess and diagnose delirium in a clinical setting~\citep{gusmao2012confusion}. We define a patient as being in delirium if the ICDSC score exceeds 3 at any point during a week-long ICU stay.

\subsubsection{Robust Model Building}

As shown in Fig.~\ref{pipeline}(d), we have designed a pipeline capable of handling the complexity of time-series data by abstracting sequential information using features' distributions. This transformation allows for the incorporation of new derived features into various machine learning models, as shown in Section~\ref{method}. Our pipeline is designed to facilitate the early integration of multimodal data sources, enhancing the model's ability to capture the dynamics of clinical contexts. To rigorously evaluate the model's performance, we implement ten-fold cross-validation ensuring a robust assessment of the model's predictive capabilities across diverse data partitions (i.e. train/test splits).

\section{Supplementary Results and Intermediate Findings}

\subsection{Clustering in the Raw Feature Space}
In this section, we present the results obtained by applying various dimensionality reduction and clustering algorithms, such as t-distributed Stochastic Neighbor Embedding (t-SNE)~\citep{tsne}, Uniform Manifold Approximation and Projection (UMAP)~\citep{mcinnes2018umap}, Principal Component Analysis (PCA)~\citep{pca}, and Independent Component Analysis (ICA)~\citep{ica} to cluster the raw features and visualize the underlying structure of the data.
\begin{figure}[ht!]
\centering
\floatconts
  {appendix_raw}
  {\caption{Visualization of clustering raw features with t-SNE, UMAP, PCA, and ICA.}}
  {\includegraphics[trim={0cm 1cm 0cm 0cm}, width=\linewidth]{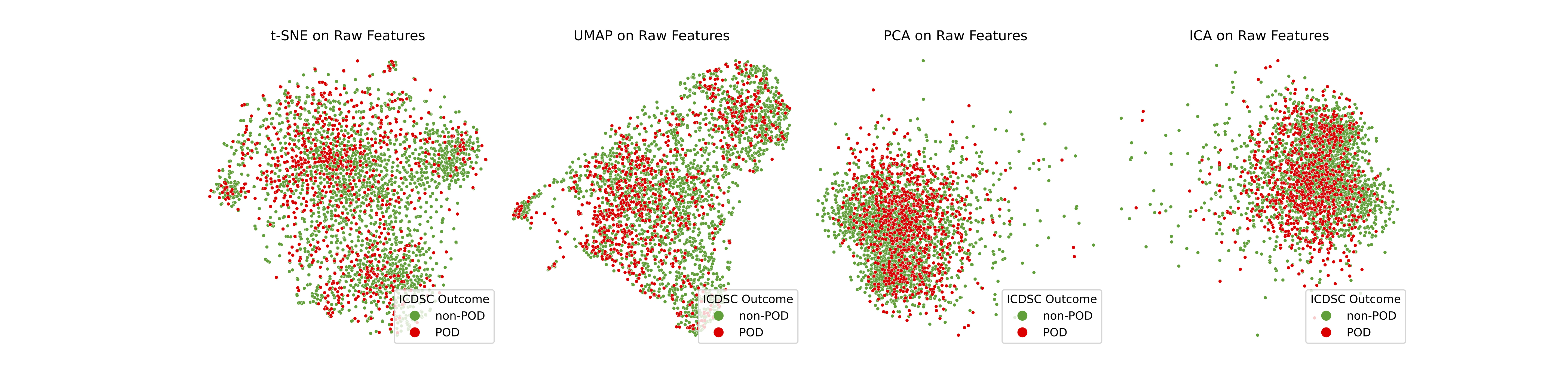}}
\end{figure}

\subsection{Risk Stratification through Different Stages}

After training our prediction models, we utilized them to calculate the individual risk probabilities (risk scores) for our patient cohort at different stages of their hospital stay. Our objective was to closely observe the evolution of patient state (that is, status) throughout the pre-operative, intra-operative, and post-operative stages, aiming to understand the progression of disease development, particularly focusing on the onset and progression of POD. We selected the best-performing model, Gradient Boosting (GB), to assess the change in the risk scores of the cohort over time. Fig.~\ref{stratification} illustrates how cohorts with different risk levels evolve across hospital stages. A major trend observed is that, prior to the onset of POD, patients predicted to be at medium or high risk are more likely to develop POD, whereas those in the low-risk cohort are less likely to experience POD. This indicates that our model is capable of detecting the potential for POD during the perioperative phase with a degree of accuracy. By computing these individual risk probabilities, we were able to chart the trajectory of each patient's risk level, providing a detailed view of how their likelihood of developing delirium changed from the time of hospital admission, through surgery, and into the critical post-operative period.

\begin{figure*}[ht!]
\floatconts
  {stratification}
  {\caption{Delirium risk stratification aross different hospital stages.}}
  {\includegraphics[trim={0cm 0cm 15cm 0cm},clip,width=0.6\linewidth]{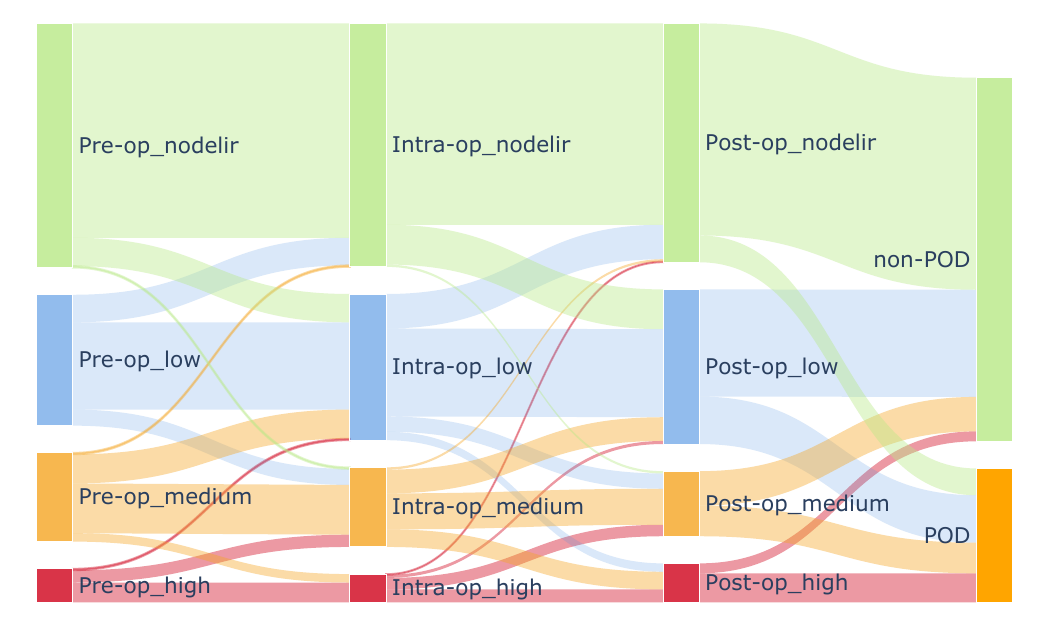}}
\end{figure*}

\subsection{Phenotype Clustering for Different Hospital Stages}

In this section, we show intermediate results of phenotype clustering. As shown in Fig.~\ref{fig:phenotypes_cumu} and~\ref{fig:phenotypes_inden}, we apply the same algorithm and pipeline to different cumulative and independent hospital stages.

\begin{figure*}[ht]
\footnotesize
\floatconts
  {fig:phenotypes_cumu}
  {\caption{Phenotype Clustering for Different Hospital Stages with Different Numbers of Clusters (cumulative cases)}}
  {%
    \subfigure[Pre$^+$-OP phenotype Clustering]{%
      \includegraphics[trim={6cm 0cm 2cm 0cm}, width=0.9\linewidth]{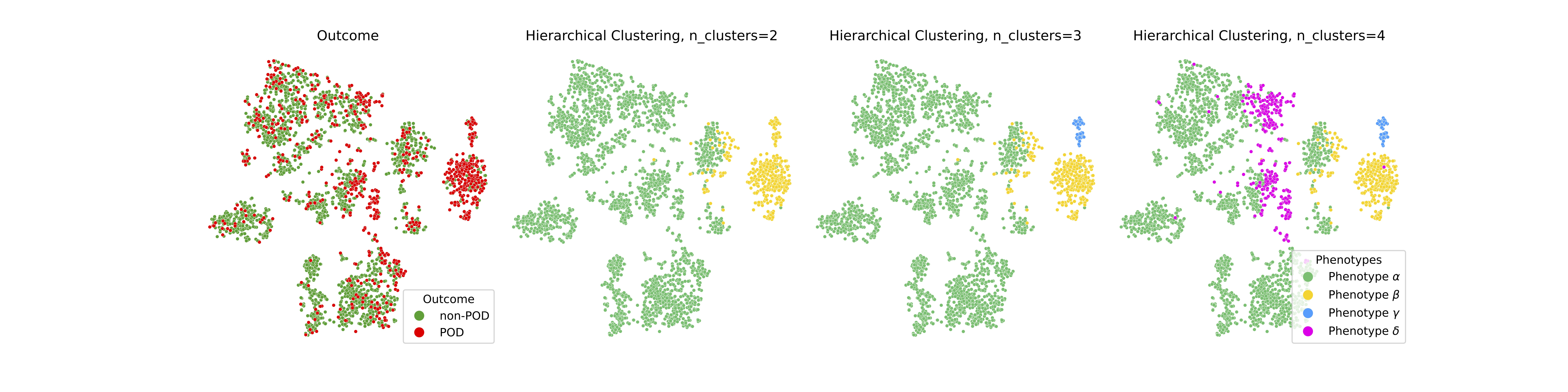}}
      \\
    \subfigure[Intra$^+$-OP phenotype Clustering]{%
      \includegraphics[trim={6cm 0cm 2cm 0cm}, width=0.9\linewidth]{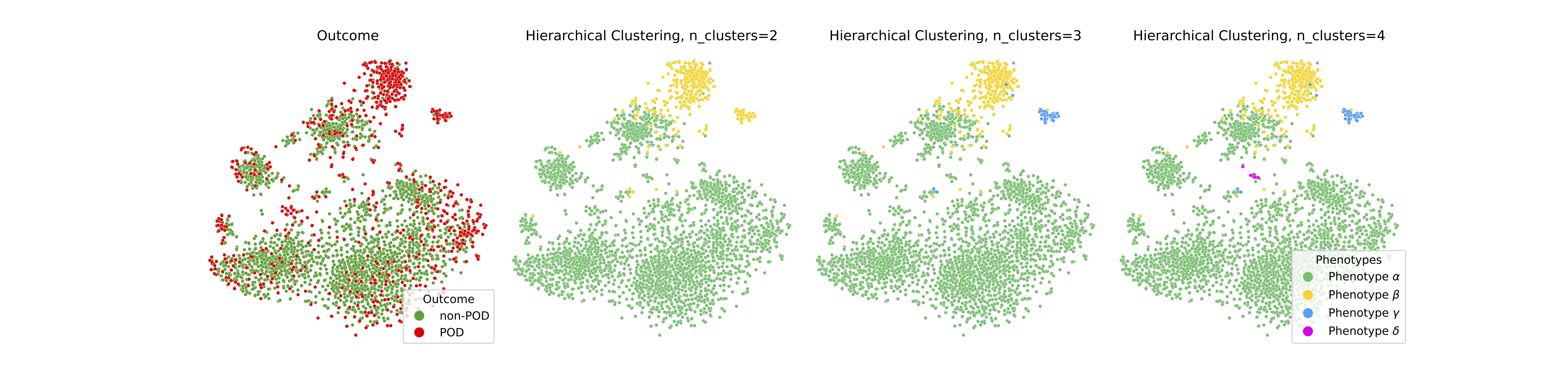}}
      \\
    \subfigure[Post$^+$-OP phenotype Clustering]{%
      \includegraphics[trim={6cm 0cm 2cm 0cm}, width=0.9\linewidth]{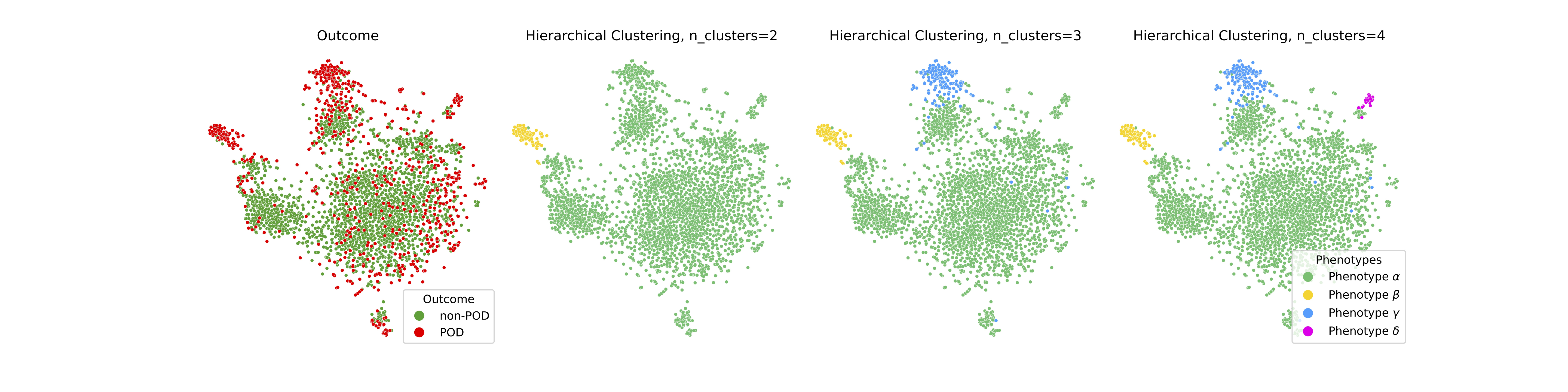}}
}
\end{figure*}

\begin{figure*}[ht]
\footnotesize
\floatconts
  {fig:phenotypes_inden}
  {\caption{Phenotype Clustering for Different Hospital Stages with Different Numbers of Clusters (independent cases)}}
  {%
    \subfigure[Pre-OP phenotype Clustering]{%
      \includegraphics[trim={6cm 0cm 2cm 0cm}, width=0.9\linewidth]{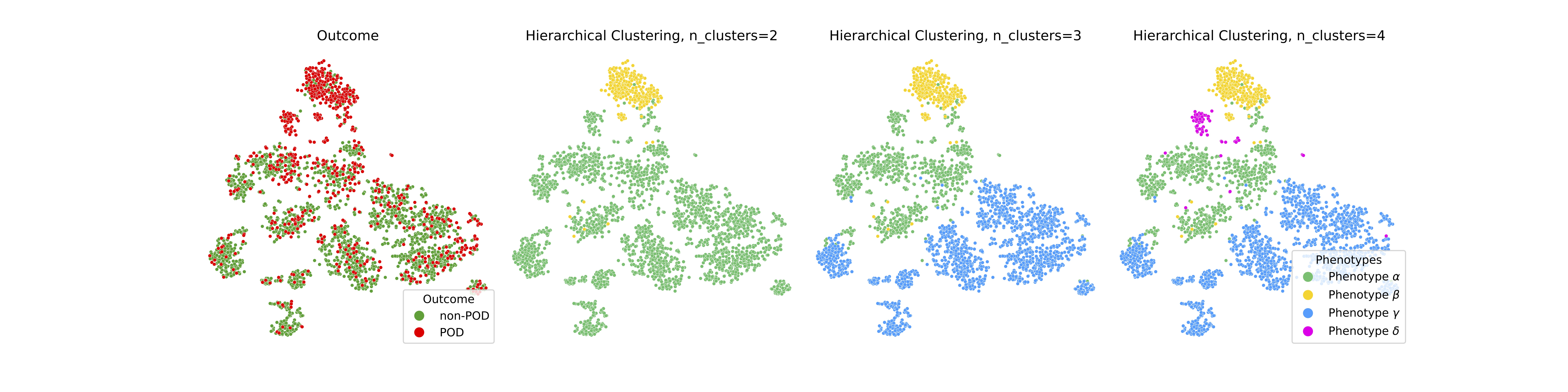}}%
      \\
    \subfigure[Intra-OP phenotype Clustering]{%
      \includegraphics[trim={6cm 0cm 2cm 0cm}, width=0.9\linewidth]{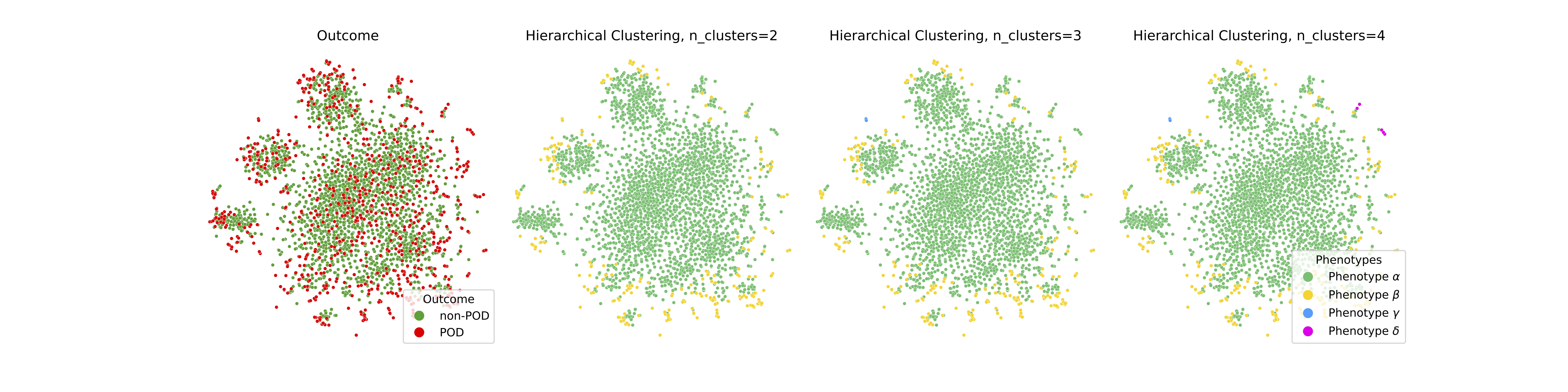}}
      \\
    \subfigure[Post-OP phenotype Clustering]{%
      \includegraphics[trim={6cm 0cm 2cm 0cm}, width=0.9\linewidth]{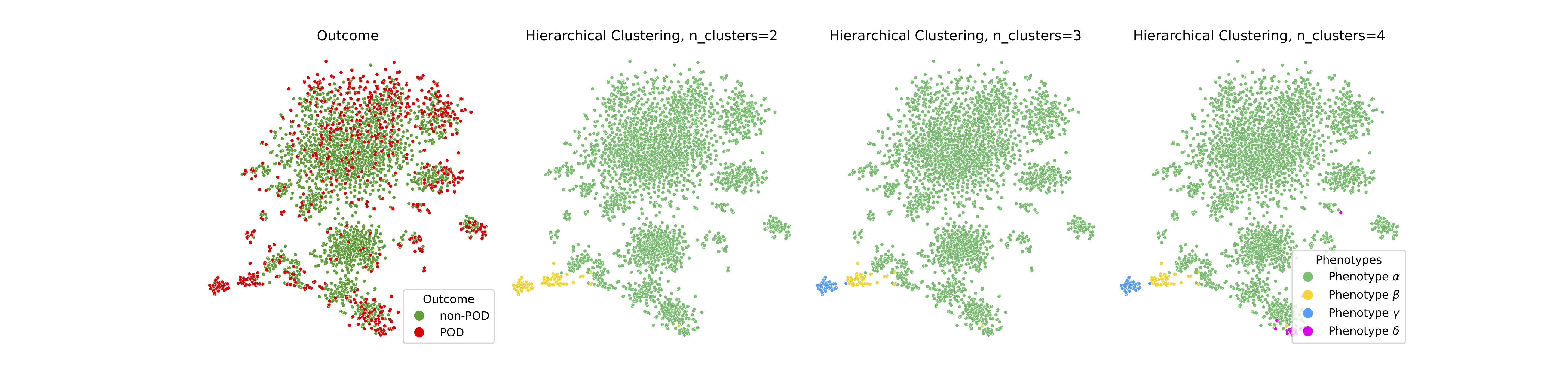}}
}
\end{figure*}

\subsection{Minimized Model}

In this section, we aim to refine these models to achieve a more minimalistic approach that maintains high accuracy while simplifying input heterogeneity. To accomplish this, we select the top $n$ most important features based on their SHAP values and train our machine learning models, as shown in Table~\ref{min}. The primary objective is to identify a minimalistic feature set for detecting postoperative delirium, which will enable healthcare professionals to focus on fewer features and enhance the usability of the model in clinical practice. By reducing the complexity of the input data while preserving the model's performance, we strive to develop a practical and efficient tool that can be seamlessly integrated into the clinical decision-making process.

\begin{table*}[ht]
\centering
\caption{The auroc for different feature settings.}
\resizebox{\textwidth}{!}{%
\begin{tabular}{c|c|cccccccc}
\toprule
\multirow{2}{*}{Stage} & \multirow{2}{*}{Model} & \multicolumn{2}{c}{Top 5} & \multicolumn{2}{c}{Top 10} & \multicolumn{2}{c}{Top 20} & \multicolumn{2}{c}{Full} \\ 
\cline{3-10}
 & & AUROC & AUPRC & AUROC & AUPRC & AUROC & AUPRC & AUROC & AUPRC \\
\midrule
\rowcolor{lightgray} & Logistic Regression & $0.637 \pm 0.0307$ & $0.353 \pm 0.0212$ & $0.639 \pm 0.0314$ & $0.36 \pm 0.0233$ & $0.696 \pm 0.0251$ & $0.437 \pm 0.0144$ & $0.698 \pm 0.023$ & $0.454 \pm 0.022$ \\ 

\rowcolor{lightgray} & MLP                 & $0.655 \pm 0.0377$ & $0.391 \pm 0.0355$ & $0.644 \pm 0.036$ & $0.366 \pm 0.0322$ & $0.694 \pm 0.0288$ & $0.434 \pm 0.0263$ & $0.704 \pm 0.029$ & $0.452 \pm 0.030$\\

\rowcolor{lightgray} & Random Forest       & $0.66 \pm 0.0341$ & $0.4 \pm 0.038$ & $0.669 \pm 0.0351$ & $0.409 \pm 0.0401$ & $0.715 \pm 0.0289$ & $0.497 \pm 0.0357$ & $0.721 \pm 0.029$ & $0.499 \pm 0.037$\\

\rowcolor{lightgray} & XGBoost             & $0.606 \pm 0.0294$ & $0.35 \pm 0.0265$ & $0.613 \pm 0.0287$ & $0.355 \pm 0.0246$ & $0.679 \pm 0.0188$ & $0.463 \pm 0.0183$ & $0.675 \pm 0.014$ & $0.445 \pm 0.019$\\

\rowcolor{lightgray} \multirow{-5}{*}{pre$^+$-OP} & Gradient Boost      & $0.657 \pm 0.0324$ & $0.38 \pm 0.0253$ & $0.663 \pm 0.0353$ & $0.392 \pm 0.0328$ & $0.713 \pm 0.0239$ & $0.484 \pm 0.0226$ & $0.714 \pm 0.026$ & $0.485 \pm 0.029$\\

\midrule
& Logistic Regression & $0.661 \pm 0.028$ & $0.366 \pm 0.0169$ & $0.685 \pm 0.0261$ & $0.413 \pm 0.0188$ & $0.699 \pm 0.0246$ & $0.449 \pm 0.0175$ & $0.666 \pm 0.029$ & $0.427 \pm 0.035$\\ 

& MLP                 & $0.664 \pm 0.0319$ & $0.383 \pm 0.0311$ & $0.691 \pm 0.0304$ & $0.43 \pm 0.0271$ & $0.697 \pm 0.0231$ & $0.458 \pm 0.0195$ & $0.669 \pm 0.030$ & $0.429 \pm 0.029$\\

& Random Forest       & $0.677 \pm 0.0305$ & $0.433 \pm 0.0247$ & $0.713 \pm 0.0278$ & $0.498 \pm 0.0364$ & $0.725 \pm 0.0284$ & $0.503 \pm 0.0307$& $0.712 \pm 0.034$ & $0.474 \pm 0.036$\\

& XGBoost             & $0.631 \pm 0.0199$ &  $0.383 \pm 0.0171$ & $0.683 \pm 0.0238$ & $0.456 \pm 0.0154$ & $0.688 \pm 0.0203$ & $0.461 \pm 0.0267$ & $0.690 \pm 0.024$ & $0.454 \pm 0.033$\\

\multirow{-5}{*}{intra$^+$-OP} & Gradient Boost      & $0.687 \pm 0.0278$ & $0.435 \pm 0.0315$ & $0.719 \pm 0.024$ & $0.487 \pm 0.0228$ & $0.721 \pm 0.0251$ & $0.49 \pm 0.0257$ & $0.704 \pm 0.027$ & $0.466 \pm 0.029$\\

\midrule
\rowcolor{lightgray} & Logistic Regression & $0.657 \pm 0.0261$ & $0.362 \pm 0.0157$ & $0.684 \pm 0.0246$ & $0.415 \pm 0.0179$ & $0.702 \pm 0.024$ & $0.472 \pm 0.0192$ & $0.659 \pm 0.023$ & $0.428 \pm 0.033$\\ 

\rowcolor{lightgray} & Multilayer Perceptron                 & $0.676 \pm 0.042$ & $0.405 \pm 0.0549$ & $0.69 \pm 0.0329$ & $0.428 \pm 0.032$ & $0.707 \pm 0.0246$ & $0.48 \pm 0.0266$ & $0.667 \pm 0.025$ & $0.433 \pm 0.035$\\

\rowcolor{lightgray} & Random Forest       & $0.681 \pm 0.0333$ & $0.447 \pm 0.0233$ & $0.713 \pm 0.0266$ & $0.498 \pm 0.033$ & $0.752 \pm 0.0229$ & $0.534 \pm 0.0317$ & $0.724 \pm 0.031$ & $0.490 \pm 0.037$\\

\rowcolor{lightgray} & XGBoost             & $0.634 \pm 0.0235$ & $0.388 \pm 0.0209$ & $0.678 \pm 0.0157$ & $0.452 \pm 0.0167$ & $0.732 \pm 0.0159$ & $0.521 \pm 0.0209$ & $0.731 \pm 0.021$ & $0.512 \pm 0.028$\\

\rowcolor{lightgray} \multirow{-5}{*}{post$^+$-OP} & Gradient Boosting      & $0.689 \pm 0.0335$ & $0.445 \pm 0.0301$ & $0.720 \pm 0.0228$ & $0.494 \pm 0.0308$ & $0.761 \pm 0.0203$ & $0.551 \pm 0.0248$ & $0.743 \pm 0.019$ & $0.533 \pm 0.029$\\
\bottomrule
\end{tabular}
}

\label{min}
\end{table*}


\clearpage
\subsection{Different cases regarding ICU delirium}

In this section, we delve into patient records to categorize them into distinct cohorts, aiming to assess the reliability of our delirium labeling process. Utilizing both the Intensive Care Delirium Screening Checklist (ICDSC) during ICU stays and the International Classification of Diseases (ICD) codes assigned post-hospital discharge, we aim to dissect the intricacies of delirium diagnosis. The ICDSC is employed by healthcare professionals in the ICU to screen for delirium symptoms systematically, while ICD codes offer a global standard for recording diagnoses and health conditions post-discharge.

Patients are categorized into four groups based on the presence or absence of these indicators:
\begin{itemize}
    \item \textbf{Confirmed ICU Delirium}: Patients with positive indicators from both ICDSC and ICD codes are classified as having confirmed ICU delirium.
    \item \textbf{ICU Delirious Symptoms}: Patients showing positive ICDSC results but without corresponding positive ICD codes are considered to exhibit delirium symptoms specifically in the ICU, suggesting potential underdiagnosis post-discharge.
    \item \textbf{Ward Delirium}: Patients with negative ICDSC but positive ICD codes are identified as experiencing delirium recognized outside the ICU, possibly indicating delirium that developed or was diagnosed later in the ward.
    \item \textbf{No Delirium}: Patients lacking positive results from both ICDSC and ICD codes are categorized as not having experienced delirium.
\end{itemize}

As shown in Fig~\ref{differentlabels}, different assessment procedures of labeling delirium highlights the intricate and possibly imprecise nature of these labels, which mirrors the challenges faced by healthcare professionals in accurately diagnosing and documenting delirium throughout various stages of patient care. The findings from this analysis emphasize the need for addressing the issue of \textbf{exsiting noise in delirium labels} when training a machine learning model. Moreover, these insights underscore the significance of refining and improving our current methods for detecting and categorizing delirium to ensure more precise and reliable diagnoses in the future.

\begin{figure*}[htbp]
\centering
\floatconts
  {differentlabels}
  {\caption{Development of subgroups through different hospital stages}}
  {\includegraphics[trim={6cm 1cm 4cm 0cm}, width=0.9\linewidth]{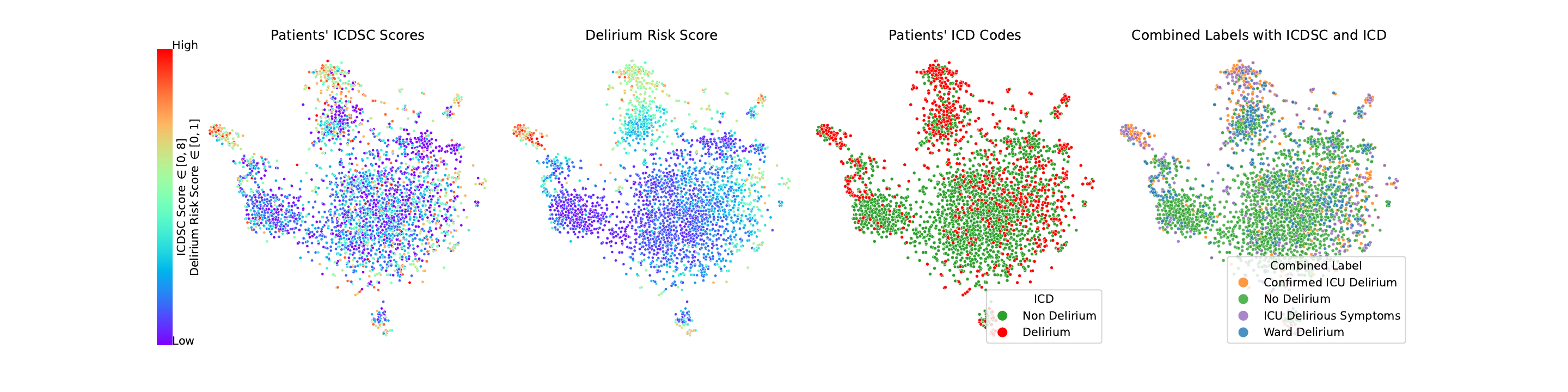}}
\end{figure*}

\end{document}